\documentclass[10pt,twocolumn,letterpaper]{article}

\usepackage[pagenumbers]{cvpr}
\usepackage{multirow}
\usepackage{graphicx}

\definecolor{cvprblue}{rgb}{0.21,0.49,0.74}
\usepackage[pagebackref,breaklinks,colorlinks,allcolors=cvprblue]{hyperref}

\title{DI-PCG: Diffusion-based Efficient Inverse Procedural Content Generation \\ for High-quality 3D Asset Creation}

\author{
Wang Zhao$^{1}$ \quad
Yan-Pei Cao$^{2}$ \quad
Jiale Xu$^{1}$ \quad
Yuejiang Dong$^{1,3}$ \quad
Ying Shan$^{1}$ \\
$^1$ ARC Lab, Tencent PCG\quad
$^2$ VAST\quad
$^3$ Tsinghua University \\
\url{https://thuzhaowang.github.io/projects/DI-PCG}
}

\begin{document}

\twocolumn[{%
\renewcommand\twocolumn[1][]{#1}%
\maketitle
\begin{center}
    \centering
    \vspace{-10pt}
    \captionsetup{type=figure}
    \includegraphics[width=1.0\linewidth]{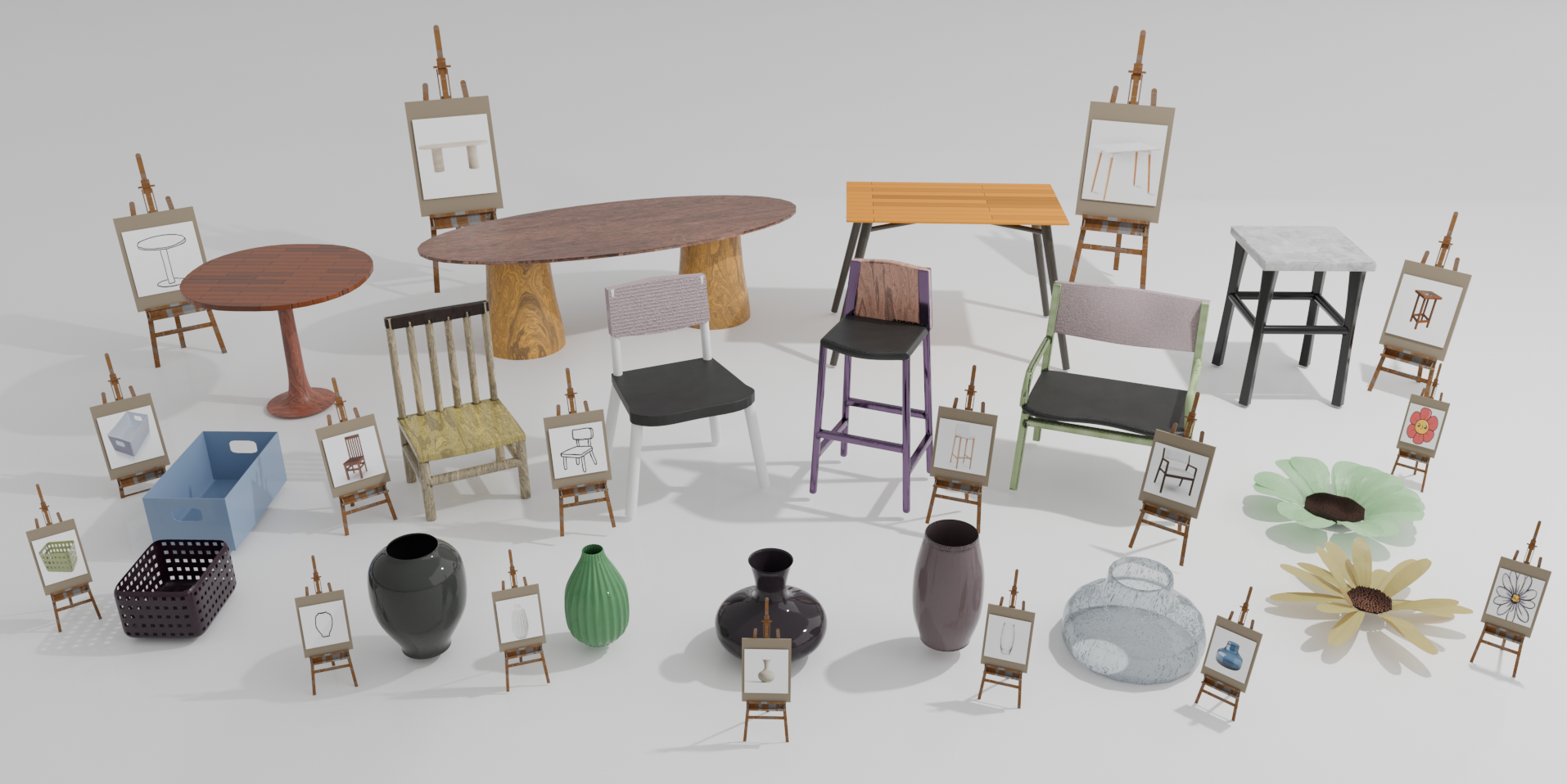}
    \captionof{figure}{Given condition images, DI-PCG can accurately estimate suitable parameters of procedural generators, resulting high fidelity 3D asset creation. Textures and materials are randomly assigned by the procedural generators for visualizations.}
    \label{fig::teaser}
\end{center}%
}]

\begin{abstract}
Procedural Content Generation (PCG) is powerful in creating high-quality 3D contents, yet controlling it to produce desired shapes is difficult and often requires extensive parameter tuning. Inverse Procedural Content Generation aims to automatically find the best parameters under the input condition. However, existing sampling-based and neural network-based methods still suffer from numerous sample iterations or limited controllability. In this work, we present DI-PCG, a novel and efficient method for Inverse PCG from general image conditions. At its core is a lightweight diffusion transformer model, where PCG parameters are directly treated as the denoising target and the observed images as conditions to control parameter generation. DI-PCG is efficient and effective. With only 7.6M network parameters and 30 GPU hours to train, it demonstrates superior performance in recovering parameters accurately, and generalizing well to in-the-wild images. Quantitative and qualitative experiment results validate the effectiveness of DI-PCG in inverse PCG and image-to-3D generation tasks. DI-PCG offers a promising approach for efficient inverse PCG and represents a valuable exploration step towards a 3D generation path that models how to construct a 3D asset using parametric models.

\end{abstract}    
\section{Introduction}
\label{sec:intro}
Procedural Content Generation (PCG) is a powerful mean to design and generate high-quality 3D contents, via algorithmic programs and rules, and has a wide application in the gaming and movie industry. Over decades, a number of works were proposed to automatically generate various 3D contents such as tree~\cite{mvech1996visual, smith1984plants, prusinkiewicz1994synthetic}, terrain~\cite{fournier1982computer, galin2019review}, building~\cite{muller2006procedural}, material~\cite{guo2020bayesian, hu2022inverse}, city~\cite{parish2001procedural, zhang2024cityx}, or even the whole natural world~\cite{raistrick2023infinite}, through different domain-specific language grammars like L-system~\cite{lindenmayer1968mathematical, prusinkiewicz1986graphical, prusinkiewicz2012algorithmic}, shape and split program~\cite{stiny1971shape}, Blender geometry nodes~\cite{blender}, etc. However, even exhibited with explicit parameter definitions, creating a desired 3D asset using PCG is highly non-trivial and requires cumbersome parameter tuning, hindering broader applications such as text or image to 3D generation.

This controlling difficulty in PCG leads to Inverse Procedural Content Generation (I-PCG), which aims to inverse the PCG task, i.e. automatically estimate the best-fit parameters from the given observations. The observations could be image, 3D, or other constraints. Similar to other non-linear and non-differential inverse problems, probabilistic sampling-based method is the golden rule for inverse PCG, where a set of samples are conducted and scored to approximate the posterior distribution given the observation. Markov chain Monte Carlo (MCMC)~\cite{metropolis1953equation, hastings1970monte} is one of the most representative methods. Many variants of MCMC~\cite{green1995reversible, talton2011metropolis, ritchie2015controlling} and different likelihood evaluation metrics~\cite{vanegas2012inverse, stava2014inverse} are explored to improve sampling efficiency and approximation accuracy. Unfortunately, most of the sampling-based methods still entail hundreds or thousands of iterations, with procedural generator forward and evaluation in each iteration, resulting a long time to finish the inverse. The key reason is that sampling-based methods do not have any data priors about the target distributions, thus need to approximate it from scratch with numerous samples. Motivated by this, several works~\cite{huang2016shape, nishida2016interactive, ritchie2016neurally, guo2020inverse, pearl2022geocode, zhou2023deeptree, zuffi2024awol} aim to utilize deep neural networks to learn the distribution correspondence between PCG parameters and input observations. Despite impressive inverse performance on certain input conditions (e.g. sketch) or categories, these methods often suffer from limited condition ability, poor generalization on real-world data, and specific designs for certain object categories, preventing their usage as a general way for inverse PCG and 3D generation.

In this work, we present DI-PCG, an innovative diffusion model based method for efficient inverse PCG from general image conditions. At its core is a light-weight diffusion transformer model, where the PCG parameters are directly treated as the denoising target and the observed image serves as the condition to control the parameter generation. Through iterative denoising score-matching training, the diffusion model learns to fit the parameter space of the current procedural generator, and can perform efficient sampling on the target posterior distribution of PCG parameters within several seconds, controlled by the condition image. The sampled parameters are then fed into PCG, resulting in high-quality 3D asset generation from images.

Our proposed DI-PCG is efficient and effective. It requires only 7.6M network parameters, 30 GPU hours to train, and several seconds to draw a sample from, thus suitable for resource-constraint scenarios. Besides efficiency, DI-PCG could effectively fit the procedural generator's parameter space, recover the corresponding parameters accurately, and generalize well to in-the-wild images, thanks to the adoption of visual foundation model features for image condition. Moreover, DI-PCG is self-contained, which only relies on current procedural generator to generate data for training, without any external data collection efforts, yet generalizes well to real-world unseen data. Both quantitative and qualitative experiments clearly verify the effectiveness of DI-PCG on inverse PCG and image-to-3D generation tasks. Figure~\ref{fig::teaser} shows some examples. The generated 3D assets are in high-quality, consistent with their condition images, and ready to use for downstream applications.

DI-PCG demonstrates a promising way of utilizing the diffusion model to learn distribution priors for efficient inverse PCG. Compared to previous sampling-based or feedforward neural network-based methods, DI-PCG features significant speed-ups and nice generalization ability. 
From another perspective, DI-PCG leverages a procedural generator and its parameters as an explicit 3D representation, and designs a diffusion model to model its distribution, enabling high-quality, ready-to-use, and editable image-to-3D asset generation. DI-PCG represents a valuable exploration step towards an encouraging 3D generation path, where how to construct a 3D asset is modeled, with a parametric model, instead of modeling 3D object itself, and the parametric model can be inversely determined given input conditions.

\section{Related Works}
\label{sec:related}

\subsection{Procedural Content Generation and Inverse}
Procedural Content Generation (PCG) is a long-standing research problem in computer graphics and vision community. L-systems~\cite{lindenmayer1968mathematical} were firstly proposed for biological modeling, and later extended to model geometry of plants~\cite{prusinkiewicz1994synthetic, mvech1996visual}. To concisely describe different object categories, many domain-specific languages were introduced such as shape grammar~\cite{stiny1971shape, lipp2008interactive}, split grammar~\cite{wonka2003instant, muller2006procedural} for generating trees~\cite{smith1984plants, prusinkiewicz1994synthetic}, man-made facades and buildings~\cite{schwarz2015advanced}. Beyond shapes, PCG is also widely used in generating textures~\cite{dang2024texpro} and materials~\cite{shi2020match, hu2023generating}. Utilizing powerful node graph grammar in modern commercial software like Blender~\cite{blender}, Infinigen~\cite{raistrick2023infinite} and Infinigen Indoors~\cite{raistrick2024infinigen} developed a broad collection of diverse procedural generators including objects, natural assets, and compositional scenes, greatly facilitating the synthetic data generation. Recently, inspired from the success of Large Language Models (LLMs), many works~\cite{sun20233d, yamada2024l3go, hu2024scenecraft, huang2024blenderalchemy, kulits2024re, zhou2024scenex, avetisyan2024scenescript, kodnongbua2023reparamcad, slim2024shapewalk} proposed to leverage the LLM reasoning capability to automatically design or edit procedural generators for 3D creation and interaction. While still limited in certain constrained scenarios, these works demonstrate promising attempts to employ general LLM agents with contexts to produce usable domain-specific languages for PCG.

Despite the ability of generating high-quality 3D assets with diversity, one of the major drawback of PCG is its difficulty to control. While easy to tune one or two specific parameters, it would be annoyingly complicated to find the appropriate combinations for tens of parameters to produce the desired shape. Inverse PCG is then introduced to inversely find the best fit parameters from the observations. Many works~\cite{cagan1993optimally, benevs2011guided, yu2011make, stava2014inverse, vanegas2012inverse} used Markov chain Monte Carlo (MCMC) methods to search the parameters. To better deal with multiple groups of parameters, Talton \textit{et al.}~\cite{talton2011metropolis} adopt Reversible Jump MCMC, with same spirit in ~\cite{ripperda2006reconstruction, ripperda2009evaluation}. Ritchie \textit{et al.}~\cite{ritchie2015controlling} further proposed stochastically-ordered sequential Monte Carlo to reduce the total numbers of PCG forward. Other optimization algorithms such as genetic~\cite{haubenwallner2017shapegenetics} was also studied for inverse PCG. PICO~\cite{krs2020pico} designed a procedural model with constraint optimizer for interactive controlling. To enable the continuous optimization, some works~\cite{garifullin2023single, stekovic2024pytorchgeonodes, gao2024diffcad} tried to make the PCG process differentiable and then optimize it using gradients.

With the tremendous success of neural networks for solving vision problems, a number of works have explored using a neural network to directly map input conditions to the PCG parameters. Ritchie \textit{et al.}~\cite{ritchie2016neurally} built a neural-guided procedural model, where certain ramdom parameters are predicted by the trained network. CSGNet~\cite{sharma2018csgnet} and InverseCSG~\cite{du2018inversecsg} focus on inferring parameters of Constructive Solid Geometry (CSG), which can be viewed as a special class of procedural modeling used in CAD. In ~\cite{nishida2016interactive, huang2016shape, pearl2022geocode}, procedural models are controlled via sketches, with convolutional neural networks (CNNs) learned to extract sketch features and regress parameters. Guo~\textit{et al.}~\cite{guo2020inverse} and DeepTree~\cite{zhou2023deeptree} focus on branching structures like tree and introduce specific designs to handle it. Different from these methods, our DI-PCG enables general image besides sketch as the input condition, and supports any procedural generator with nearly zero modifications of code. By leveraging the best practices from recent diffusion models, such as using pre-trained visual foundation model features of input image as condition, and transformer-based diffusion denoising architecture, DI-PCG achieves accurate and generalizable inverse results for different generators. 

\subsection{Diffusion Models for 3D Generation}
Diffusion models have achieved remarkable progress in generative modeling, with increasing popularity in 3D generation. Due to the scarcity of 3D data, early works attempted to utilize 2D diffusion priors through score distillation sampling~\cite{pooledreamfusion} and its enhancements~\cite{chen2023fantasia3d, metzer2023latent, seolet, wang2024prolificdreamer, lin2023magic3d}. This distillation inherently lacks view consistency and 3D priors, often leading to blurry textures and multi-head Janus problem. To mitigate this issue, Zero-1-to-3~\cite{liu2023zero} proposed to generate novel view images under required camera viewpoints, and reconstruct 3D representation using generated multiview images. Following this line of research, a number of works~\cite{shimvdream, liusyncdreamer, shi2023zero123++, long2024wonder3d, qiu2024richdreamer, voleti2025sv3d} explored fine-tuning 2D diffusion models to directly generate multiview images via carefully designed view interaction, which greatly improve the view consistency and thus benefit 3D generation.

With the advent of large-scale 3D datasets such as Objaverse~\cite{deitke2023objaverse}, training 3D native diffusion models is made possible. Different kinds of 3D shape representations are explored such as point cloud, voxel, mesh, implicit functions, etc. Point-E~\cite{nichol2022point} pioneered the denoising diffusion on point cloud. LION~\cite{vahdat2022lion} and SLIDE~\cite{lyu2023controllable} further introduced latent point diffusion model with point cloud VAE to enhance the compactness. 
To directly model 3D surfaces, PolyDiff~\cite{alliegro2023polydiff} represented meshes as quantized triangle soups and applied diffusion model on triangle vertex coordinates. MeshDiffusion~\cite{liu2023meshdiffusion}, in another way, utilized deformable marching tetrahedra~\cite{shen2021deep} representation for meshes and trained a diffusion model upon it. By exploiting sparse voxel hierarchy or Octree-base latent voxel representations, XCube~\cite{ren2024xcube} and OctFusion~\cite{xiong2024octfusion} managed to relief the memory-resolution trade-off of 3D voxels and train diffusion models over latent voxels, achieving detailed 3D generation results. 

Different from above explicit 3D representations, many works focus on implicit representations which features higher compression ratio, infinite decoding resolution, and intrinsic smoothness. SDFusion~\cite{cheng2023sdfusion} employed 3D VAE to decode SDF fields from denoised latent variables. 3DGen~\cite{gupta20233dgen} and Direct3D~\cite{wu2024direct3d} selected triplane~\cite{chan2022efficient} as the representation, while Michelangelo~\cite{zhao2024michelangelo}, 3DShape2VecSet~\cite{zhang20233dshape2vecset} and CLAY~\cite{zhang2024clay} adopt pure 3D shape latent vectors with VAEs to fully unleash the scaling ability.

Image conditioned inverse PCG can be viewed as image to 3D generation. From this view, DI-PCG essentially takes the procedural generator and its parameters as a powerful, highly compact, editable 3D representation and trains a diffusion model on top of it for high-quality 3D generation.  

\begin{figure*}[t]
    \centering
    \includegraphics[width=0.95\linewidth]{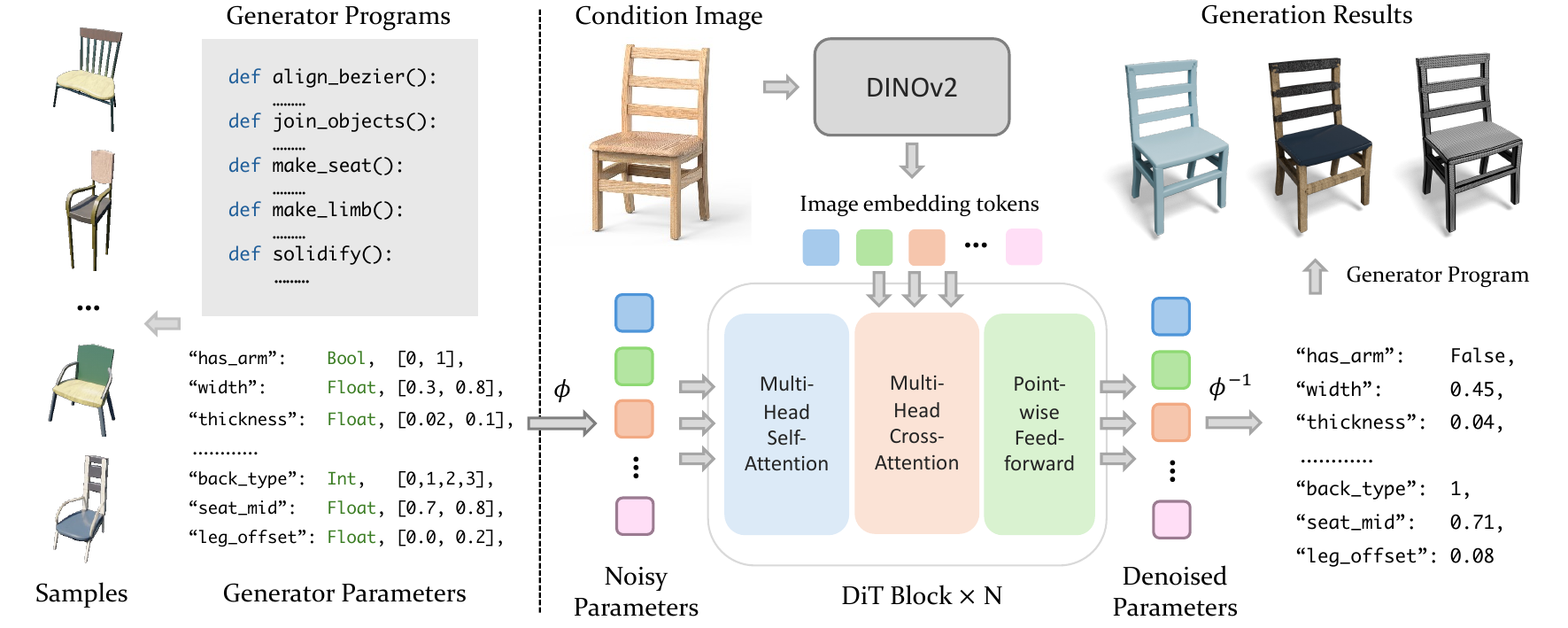}
    \caption{Overview of DI-PCG. (Left) The procedural generator consists of programs and parameters, and can be randomly sampled to produce various shapes. (Right) To control it with images, DI-PCG trains a denoising diffusion model directly upon canonicalized generator parameters, using DINOv2 to extract condition image features and inject them via cross attention. The resulting parameters are projected back to original ranges and then fed into the generator, delivering high-quality 3D generation with neat geometry and meshing.}
    \label{fig::pipeline}
    \vspace{-10pt}
\end{figure*}
\section{Methods}

\subsection{Preliminaries}
\noindent\textbf{Procedural Generator.} Procedural generator defines algorithmic rules with a set of parameters to create an asset. A generator usually handles one specific category of objects, such as a chair, vase, tree, etc. For example, a chair is procedurally constructed by the selected parameters which describe the back type of the chair, the leg height, the numbers of bars, the existence of arms, etc. Theoretically, it can generate infinitely many variants of objects by randomly sampling parameters. In practice, the capability of a generator to provide diverse instances is determined by the generality and granularity of its rules.

\noindent\textbf{Diffusion Model.} A diffusion model consists of a forward noising and reverse denoising process. The forward process gradually corrupts clean data $\mathbf{x_0}$ into a Gaussian distribution $\mathcal{N}(\mathbf{0}, \mathbf{I})$ by:
$
    q(\mathbf{x_t} | \mathbf{x_0}) = \mathcal{N} (\mathbf{x_t};\sqrt{\bar\alpha_t} \mathbf{x_0}, (1-\bar\alpha_t)\mathbf{I}),
$
where $\mathbf{x_0}$ is the input data, $t$ is the timestep and $\bar\alpha_t$ are constant hyperparameters. With the reparameterization trick, we can sample $\mathbf{x_t} = \sqrt{\bar\alpha_t} \mathbf{x_0} + \sqrt{1-\bar\alpha_t} \mathbf{\epsilon_t}$, where $\mathbf{\epsilon_t} \sim \mathcal{N}(\mathbf{0}, \mathbf{I})$. The reverse process is then defined through a Markov chain: $p_\theta(\mathbf{x_{t-1}}|\mathbf{x_t}) = \mathcal{N}(\mathbf{\mu}_\theta (\mathbf{x_t}), \mathbf{\Sigma}_\theta(\mathbf{x_t})).$  By parameterizing $\mathbf{\mu}_\theta$ as a noise prediction network $\mathbf{\epsilon}_\theta$, the reverse process is trained via the variational lower bound, with the objective reduced to the mean square error (MSE) between the predicted noise and the ground truth noise: 
\begin{equation}
\label{eq::diffusion_loss}
    \mathcal{L}_\theta = \mathbb{E}_{\mathbf{x_0},t,\mathbf{\epsilon_t}} \| \mathbf{\epsilon}_\theta(\mathbf{x_t},t) - \mathbf{\epsilon_t} \|_2^2.
\end{equation}
After training, the diffusion model can sample directly on the data distribution of $\mathbf{x}$ from a Gaussian distribution noise.

\subsection{Diffusion Model for Inverse PCG}
Our proposed DI-PCG considers the procedural generator with its parameters as a controllable 3D shape representation, and carefully designs and trains a diffusion model for the parameters, enabling to efficiently sample the target parameters under condition, as illustrated in Figure~\ref{fig::pipeline}. Next, we will describe in detail the representation, architecture, condition scheme and the data preparation process.

\begin{figure*}[tb]
    \centering
    \small
    \setlength{\tabcolsep}{0pt}
    \begin{tabular}{cccccc}

    \centering
        {\includegraphics[width=0.15\linewidth]{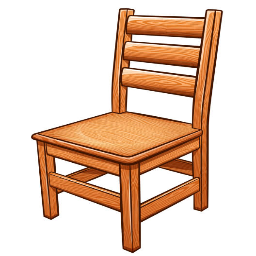}} &  
        {\includegraphics[width=0.15\linewidth]{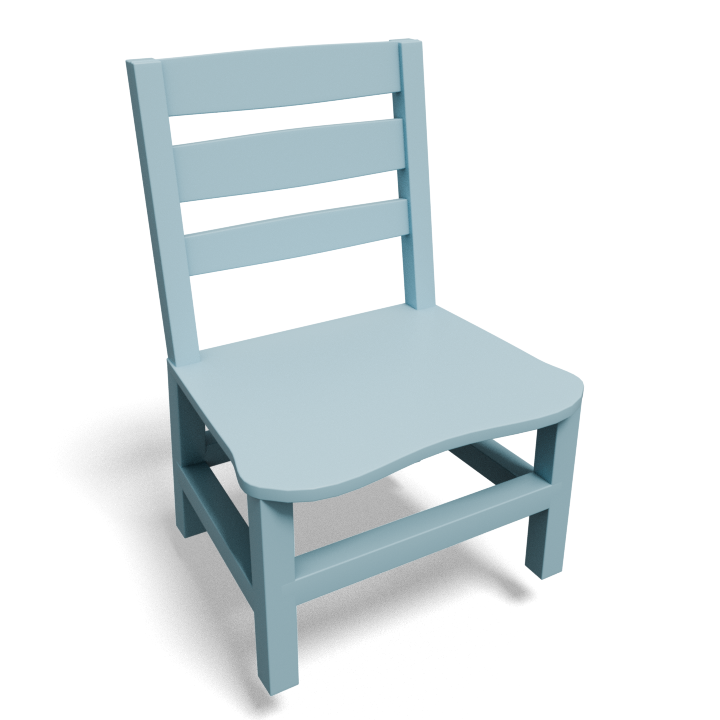}} &  
        {\includegraphics[width=0.15\linewidth]{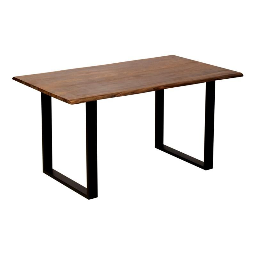}} &
        {\includegraphics[width=0.15\linewidth]{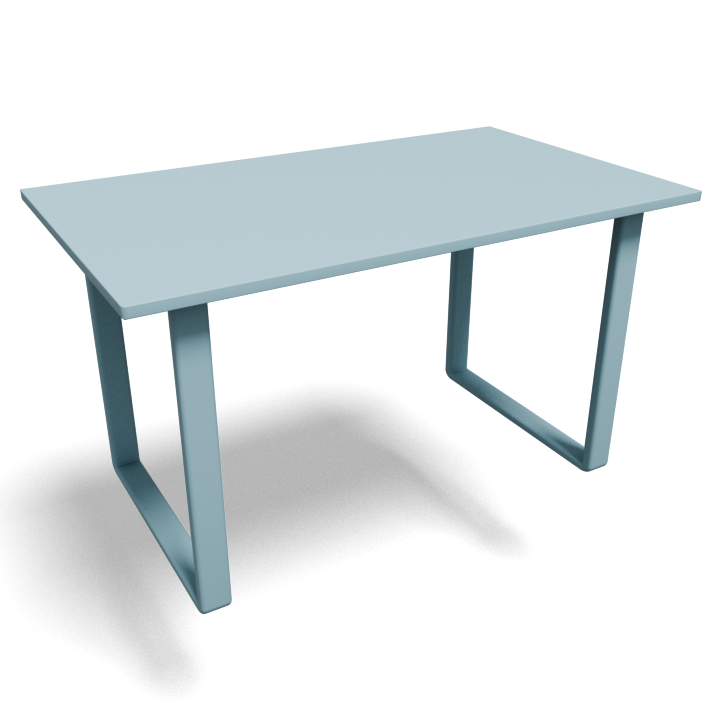}} &
        {\includegraphics[width=0.15\linewidth]{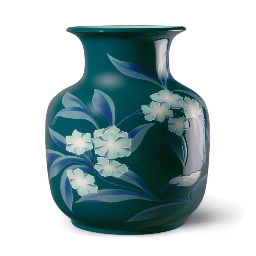}} &
        {\includegraphics[width=0.15\linewidth]{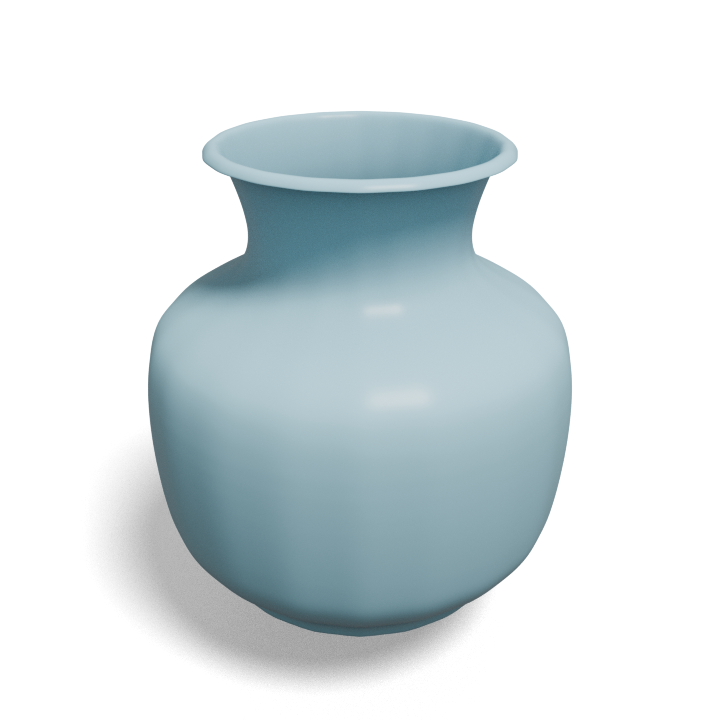}}\\

        {\includegraphics[width=0.15\linewidth]{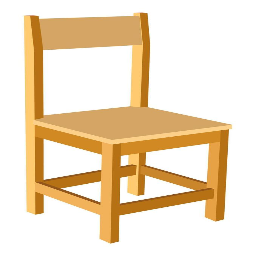}} &  
        {\includegraphics[width=0.15\linewidth]{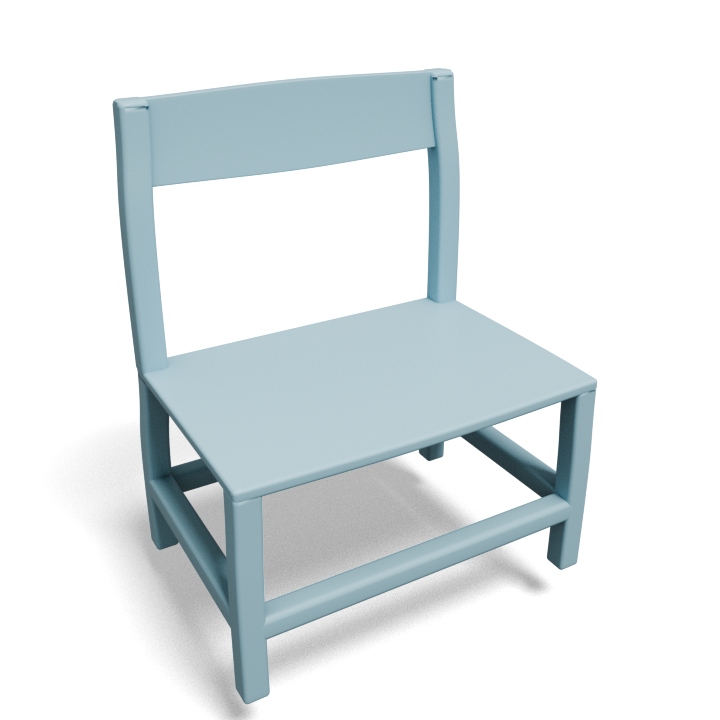}} &  
        {\includegraphics[width=0.15\linewidth]{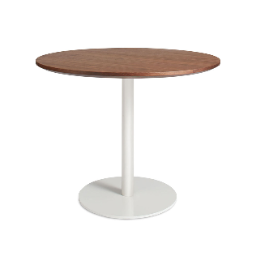}} &
        {\includegraphics[width=0.15\linewidth]{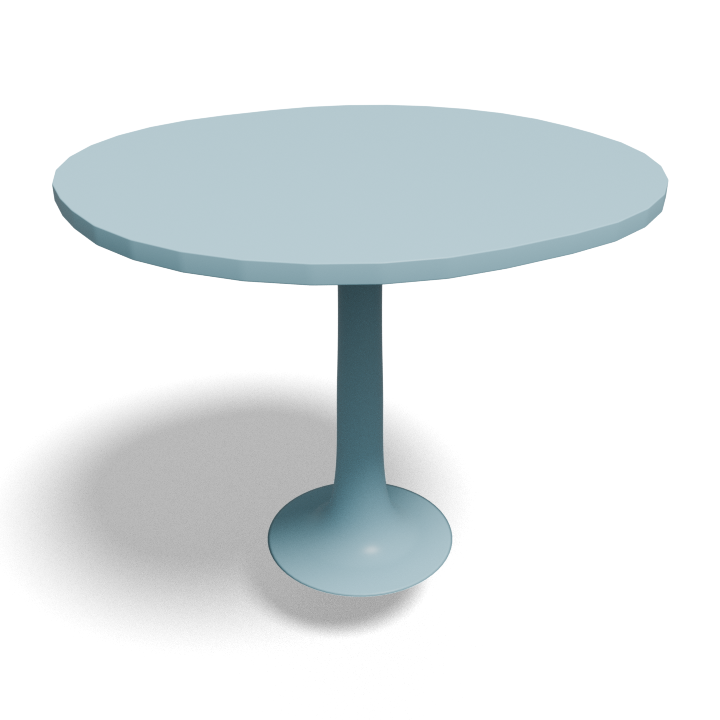}} &
        {\includegraphics[width=0.15\linewidth]{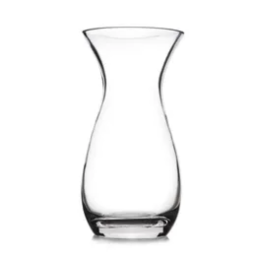}} &
        {\includegraphics[width=0.15\linewidth]{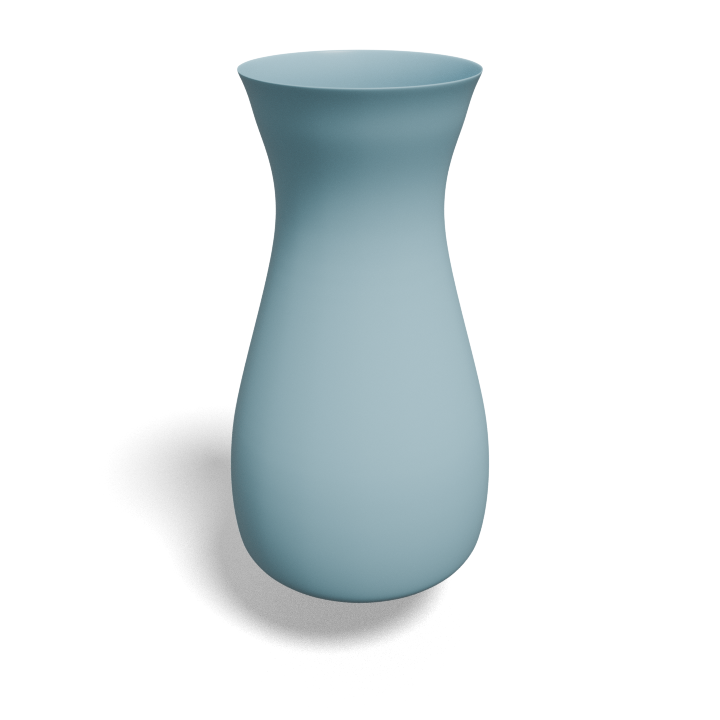}}\\

        {\includegraphics[width=0.15\linewidth]{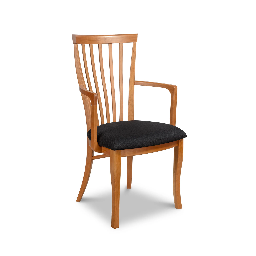}} &  
        {\includegraphics[width=0.15\linewidth]{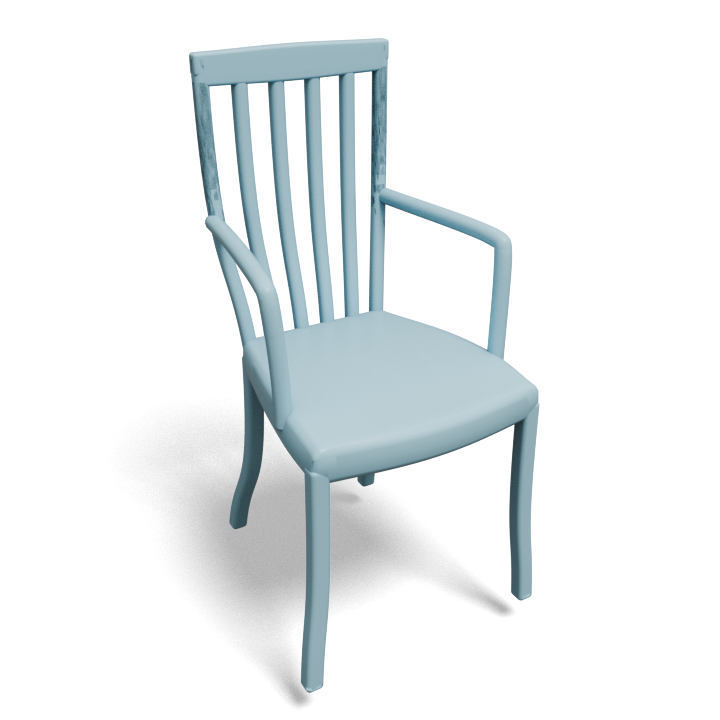}} &  
        {\includegraphics[width=0.15\linewidth]{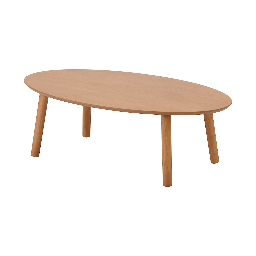}} &
        {\includegraphics[width=0.15\linewidth]{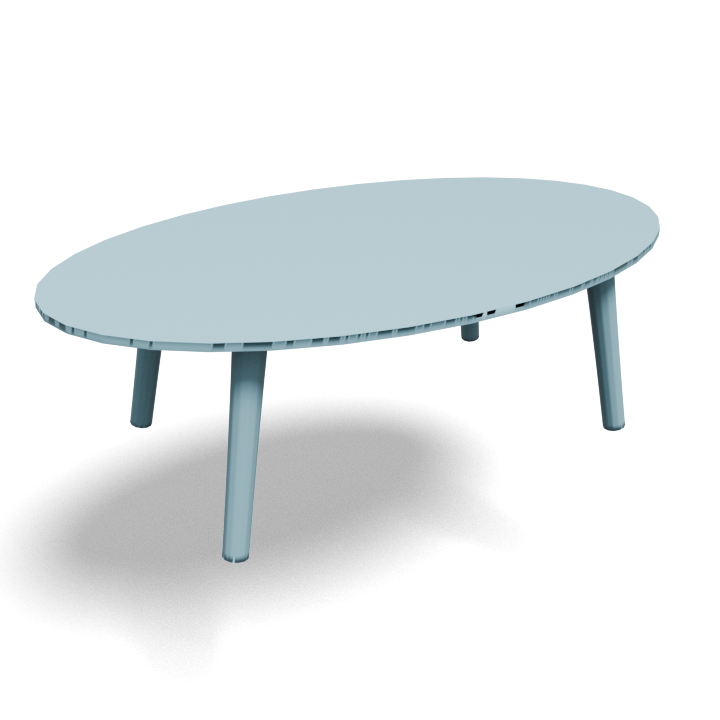}} &
        {\includegraphics[width=0.15\linewidth]{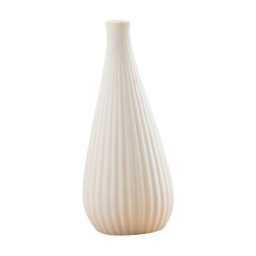}} &
        {\includegraphics[width=0.15\linewidth]{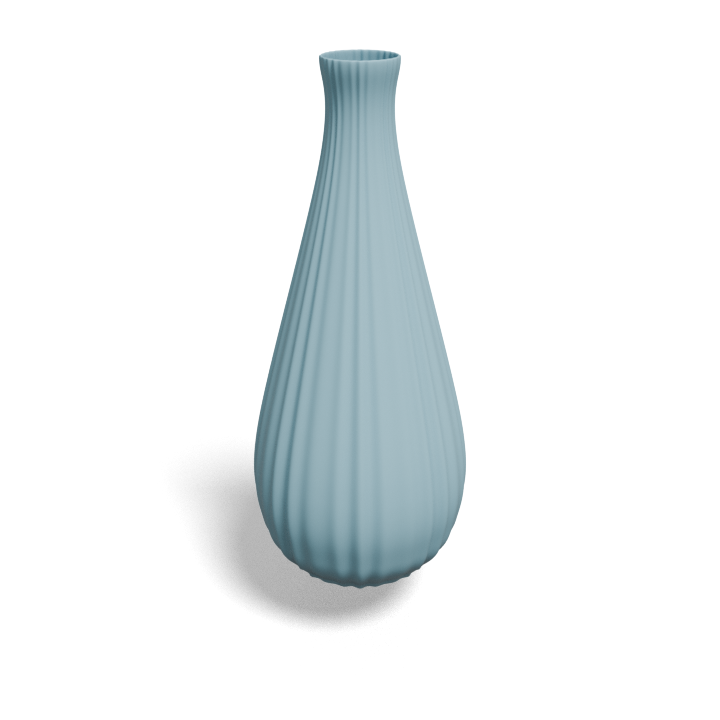}}\\


        {\includegraphics[width=0.15\linewidth]{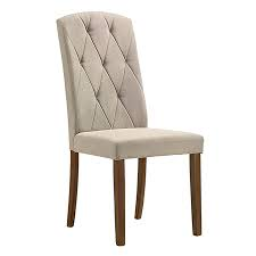}} &  
        {\includegraphics[width=0.15\linewidth]{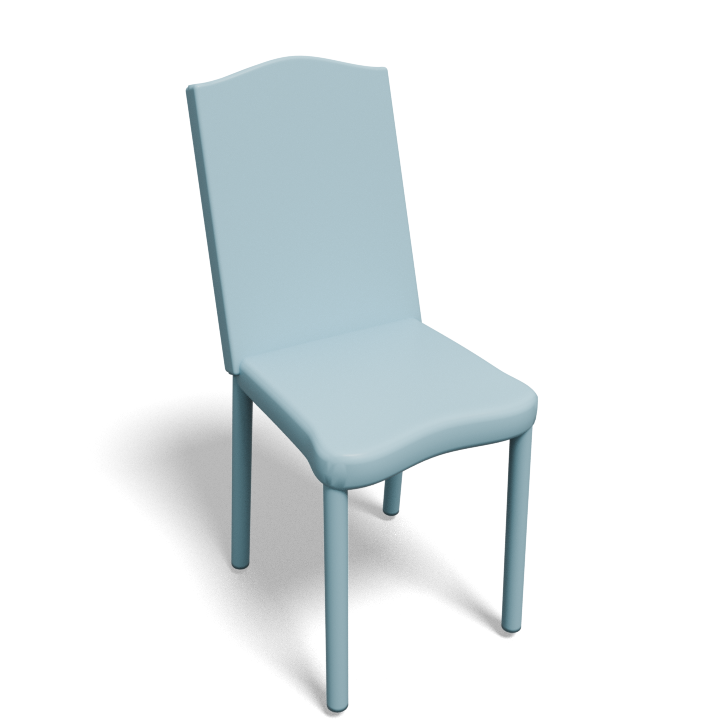}} &  
        {\includegraphics[width=0.15\linewidth]{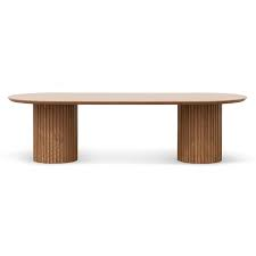}} &
        {\includegraphics[width=0.15\linewidth]{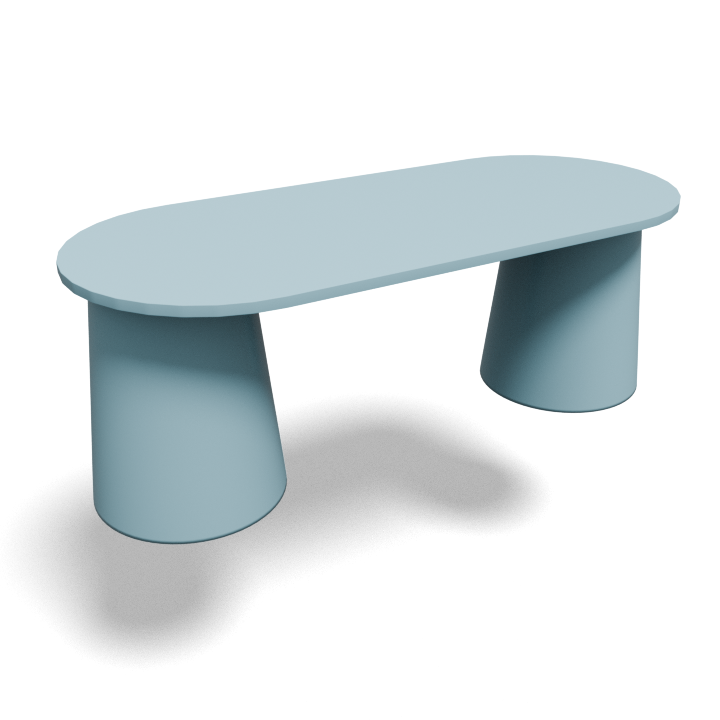}} &
        {\includegraphics[width=0.15\linewidth]{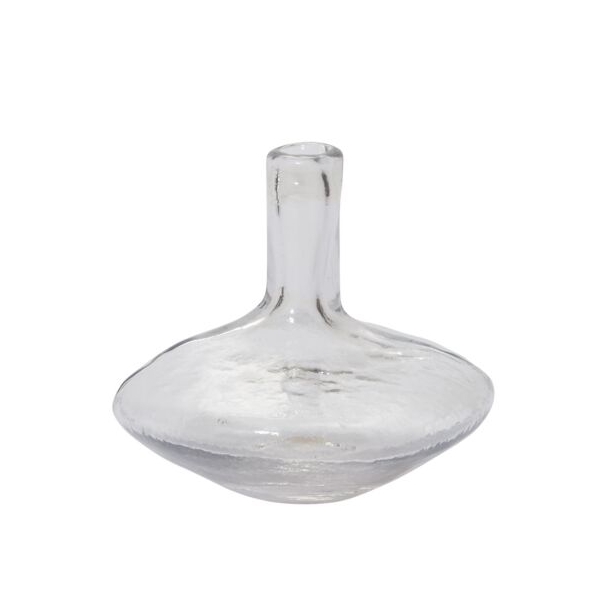}} &
        {\includegraphics[width=0.15\linewidth]{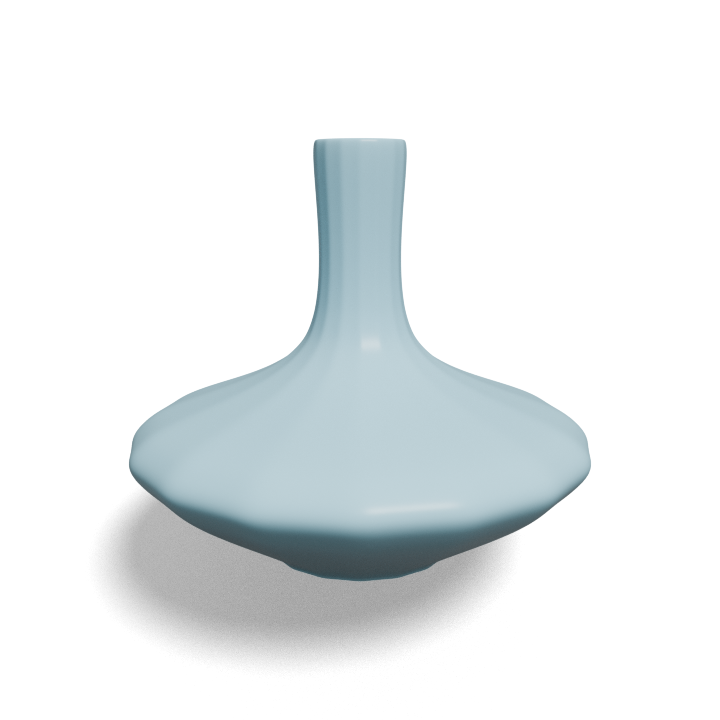}}\\

        Chair images & Results & Table images & Results & Vase images & Results \\

    \end{tabular}
    \caption{Qualitative results for chair, table, and vase generations. Input images are collected from the internet.}
    \label{fig::ipcg-results}
    \vspace{-10pt}
\end{figure*}
\noindent\textbf{Representation.} We directly treat the parameters of the procedural generator as the parametric representation of the 3D models, and learn to sample it with diffusion models. Specifically, we assume that the given procedural generator provides a list of its controllable random parameters $\mathbf{p} = \{p_0, p_1, p_2, ..., p_N\}$, and each parameter has its own sampling range, e.g. minimum and maximum values for continuous parameters, and all available choices for discrete parameters. If not provided, we manually derive them from the procedural generator's code. Since the procedural generator has both continuous and discrete parameters, which is difficult for the diffusion model to jointly model, we first make the discrete parameters continuous. We uniformly cut $[-1, 1]$ into pieces where each piece corresponds to a discrete choice. To facilitate training, the continuous parameters are also normalized to $[-1, 1]$ according to the minimum and maximum values. We denote these canonicalization operations together as a reversible projection $\mathbf{\phi}$ from the original parameter set to the normalized continuous representation $\mathbf{x} = \mathbf{\phi}(\mathbf{p})$. These normalized parameters $\mathbf{x} \in [-1,1]^{N\times 1}$ are then used in the diffusion noising and denoising process. During inference, the sampled normalized parameters are projected back to the original generator parameters using $\mathbf{p} = \mathbf{\phi}^{-1}(\mathbf{x})$, and the 3D asset is then generated via the procedural generator with $\mathbf{p}$.

\noindent\textbf{Model architecture.} Following recent successful practices~\cite{chen2023pixart, videoworldsimulators2024, zhang2024clay} in both 2D and 3D generative modeling, we employ the Diffusion Transformer (DiT)~\cite{peebles2023scalable} model. The DiT model, which served as $\epsilon_{\theta}$ in Eq.~\ref{eq::diffusion_loss}, predicts the noise at each timestep $t$ via cross and self attentions:
\begin{equation}
    \epsilon_{\theta}(\mathbf{x_t}, t, \mathbf{c}) = \{\operatorname{CrossAttn}(\operatorname{SelfAttn}(\mathbf{x_t}), \mathbf{c})\}^{L}
    \label{eq::dit}
\end{equation}
where $\mathbf{x_t}$ is the noisy version of $\mathbf{x_0}$, and $\mathbf{c}$ represents the condition features. $L$ denotes the number of attention layers. Since our procedural parameter representation is fairly expressive and compact, the denoising variable $\mathbf{x}$ usually only contains dozens of tokens. Thus, we can use a lightweight transformer model to process it. We build the DiT with 12 attention layers with 6 heads, and the hidden feature dimensions set to 192, resulting in an efficient model with 7.6M parameters. Compared to large-scale 3D generative models~\cite{zhang2024clay, li2024craftsman, wu2024direct3d} with hundreds of millions or billions of parameters for learning general objects, DI-PCG takes a different path, where a tiny, generator-specific model is responsible for creating category-specific 3D objects in high quality. With the increasing number of available procedural generators, DI-PCG can be potentially extended to a diffusion model collection, and different model combinations for various categories can be deployed to fulfill the application demands in a flexible way.

\begin{figure*}[tb]
    \centering
    \small
    \setlength{\tabcolsep}{0pt}
    \begin{tabular}{cccccc}

    \centering
        {\includegraphics[width=0.145\linewidth]{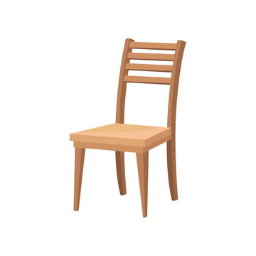}} &  
        {\includegraphics[width=0.145\linewidth]{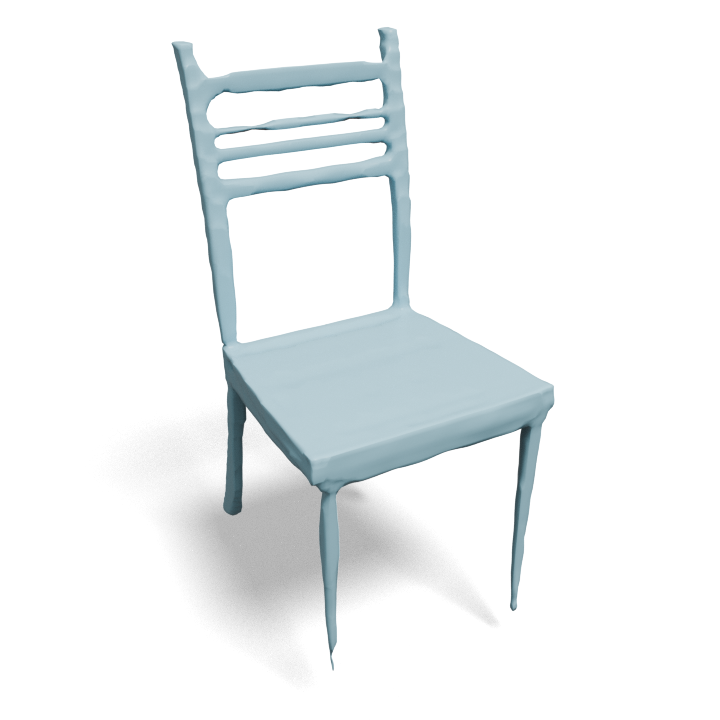}} &  
        {\includegraphics[width=0.145\linewidth]{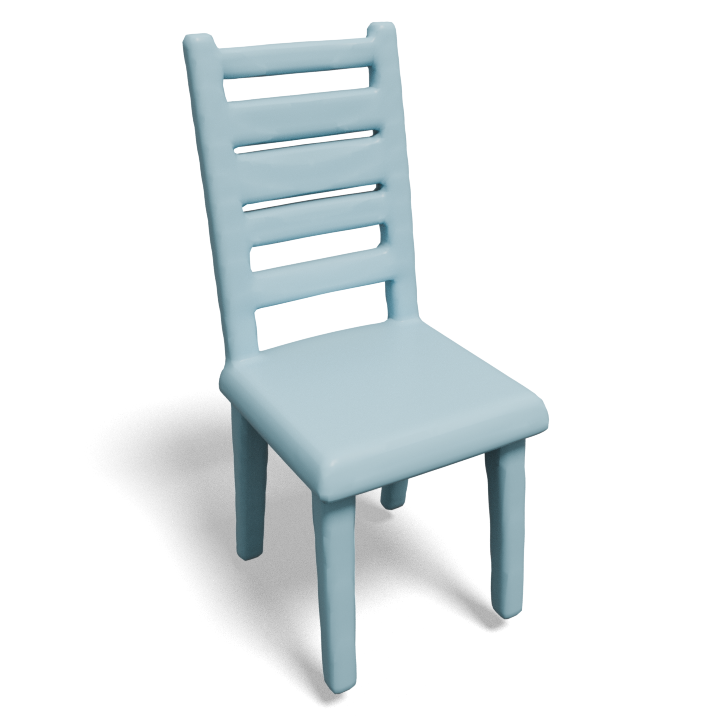}} &
        {\includegraphics[width=0.145\linewidth]{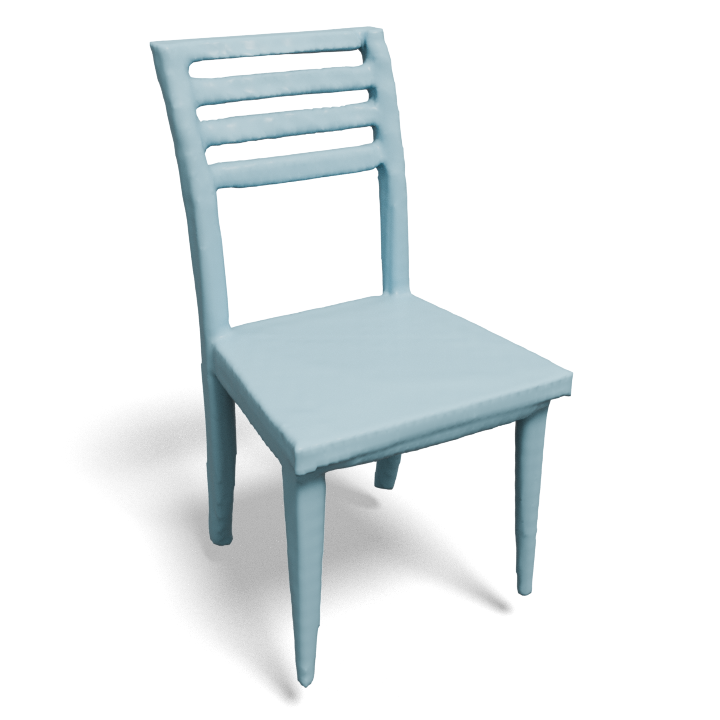}} &
        {\includegraphics[width=0.145\linewidth]{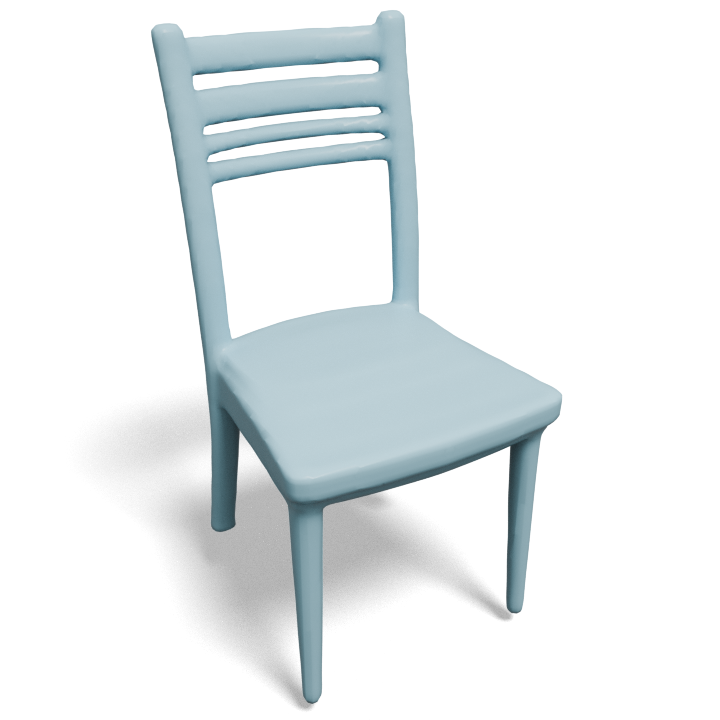}} &
        {\includegraphics[width=0.145\linewidth]{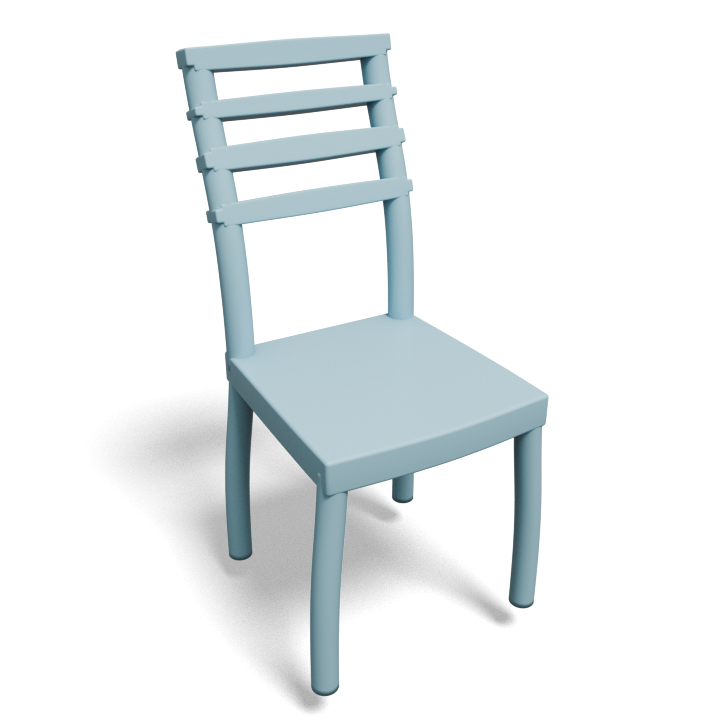}}\\

        {\includegraphics[width=0.145\linewidth]{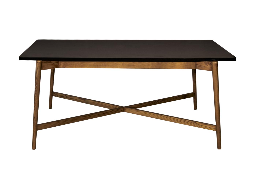}} &  
        {\includegraphics[width=0.145\linewidth]{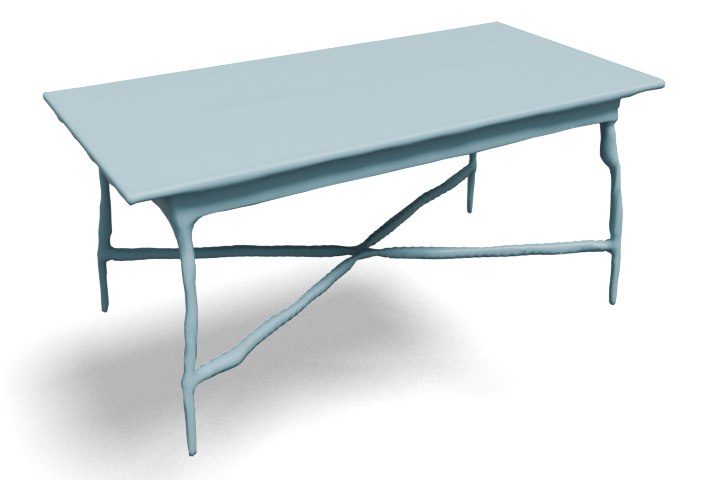}} &  
        {\includegraphics[width=0.145\linewidth]{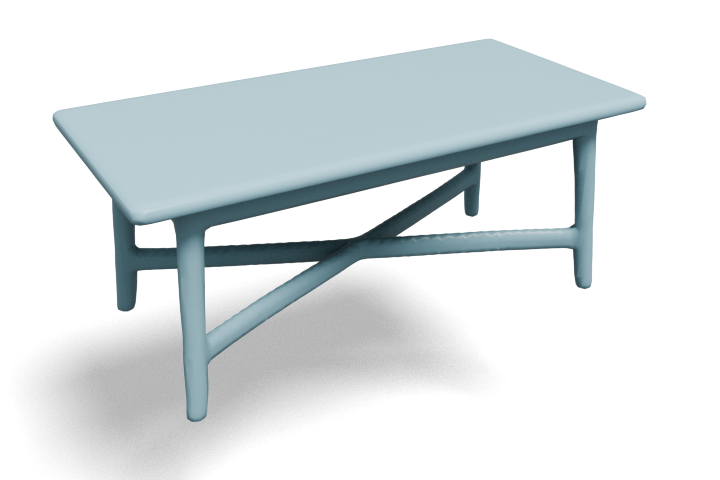}} &
        {\includegraphics[width=0.145\linewidth]{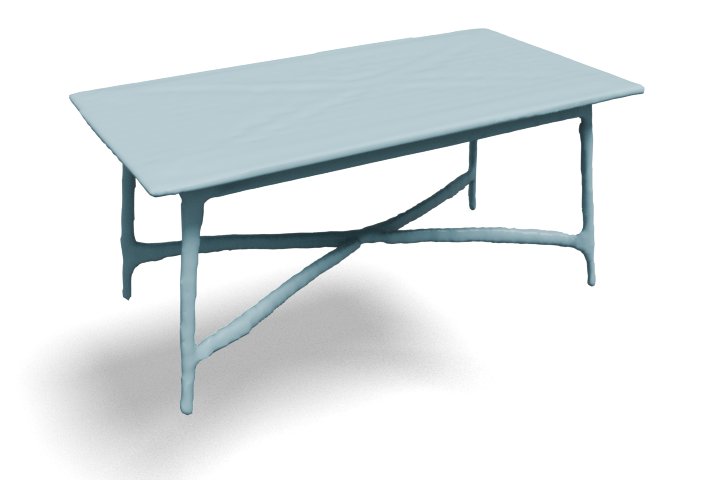}} &
        {\includegraphics[width=0.145\linewidth]{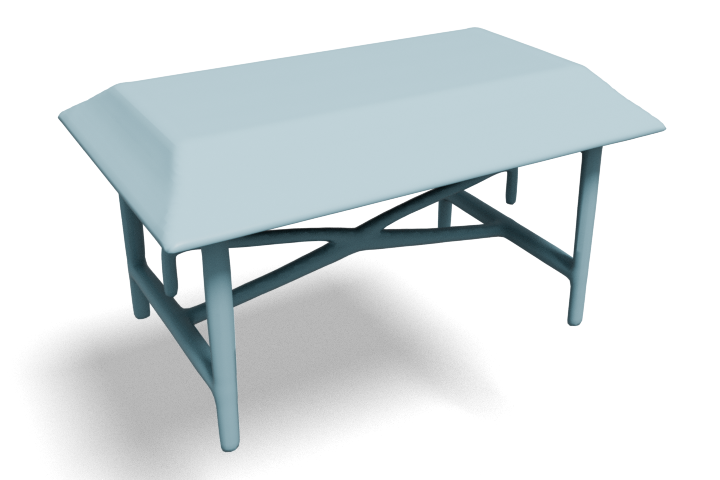}} &
        {\includegraphics[width=0.145\linewidth]{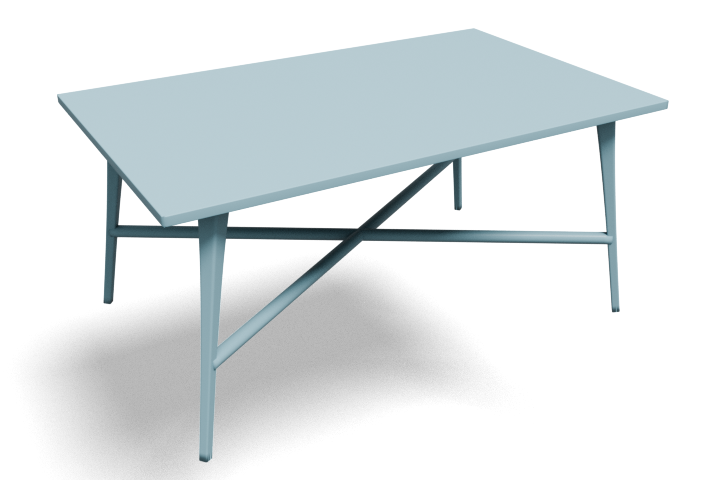}}\\

        {\includegraphics[width=0.145\linewidth]{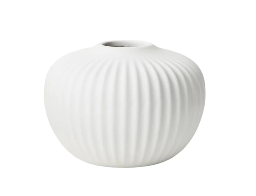}} &  
        {\includegraphics[width=0.145\linewidth]{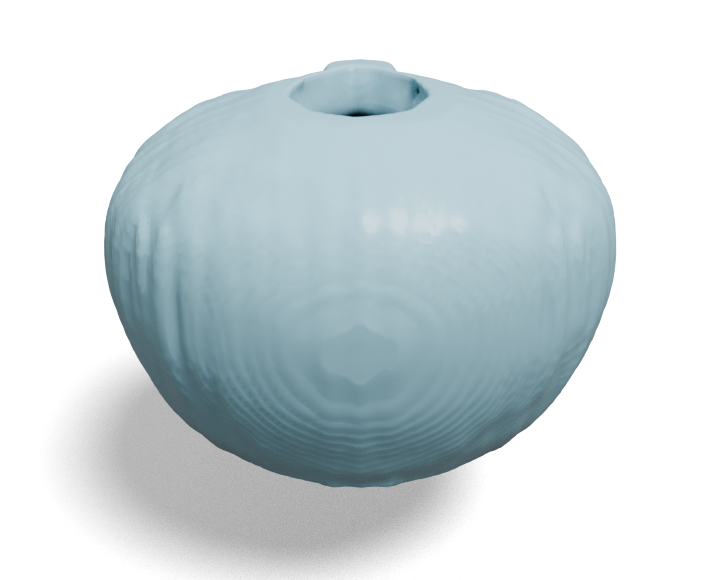}} &  
        {\includegraphics[width=0.145\linewidth]{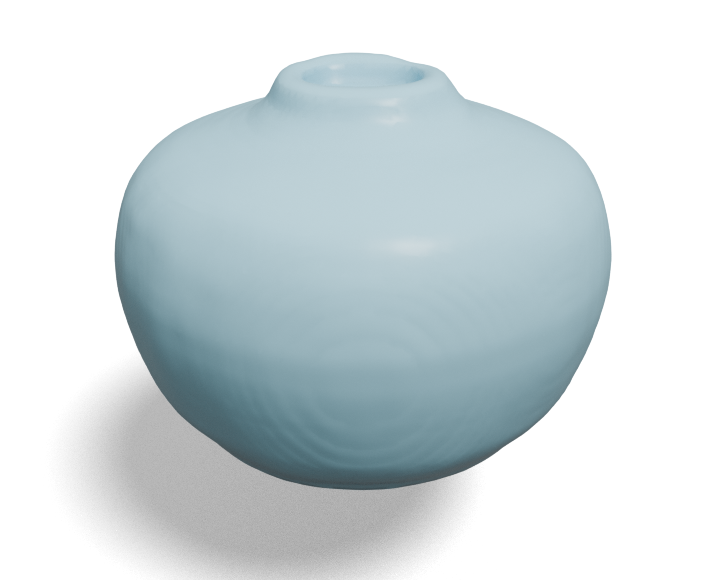}} &
        {\includegraphics[width=0.145\linewidth]{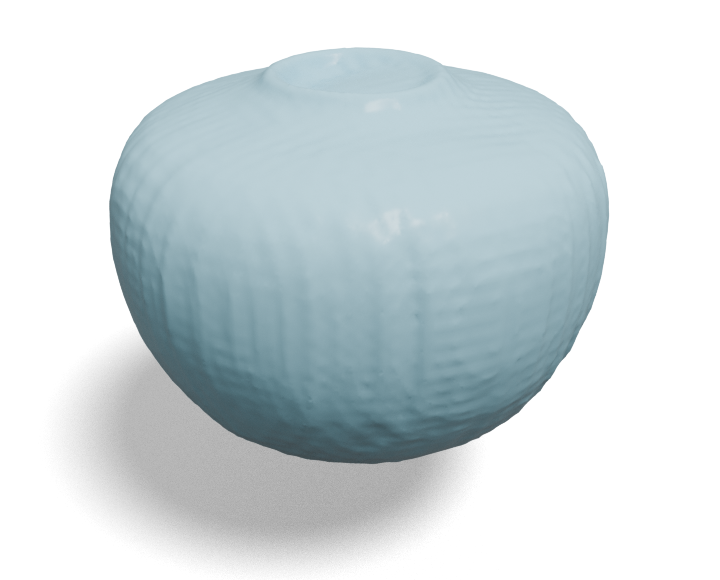}} &
        {\includegraphics[width=0.145\linewidth]{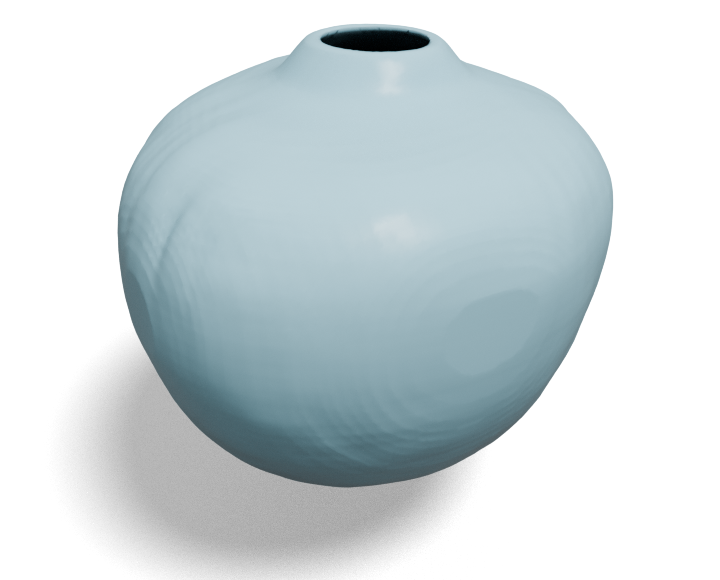}} &
        {\includegraphics[width=0.145\linewidth]{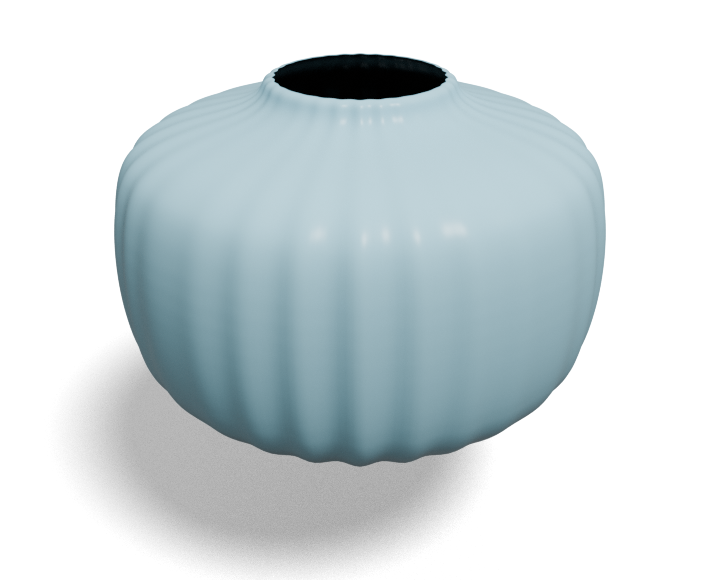}}\\

        {\includegraphics[width=0.145\linewidth]{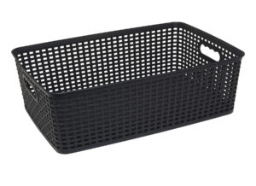}} &  
        {\includegraphics[width=0.145\linewidth]{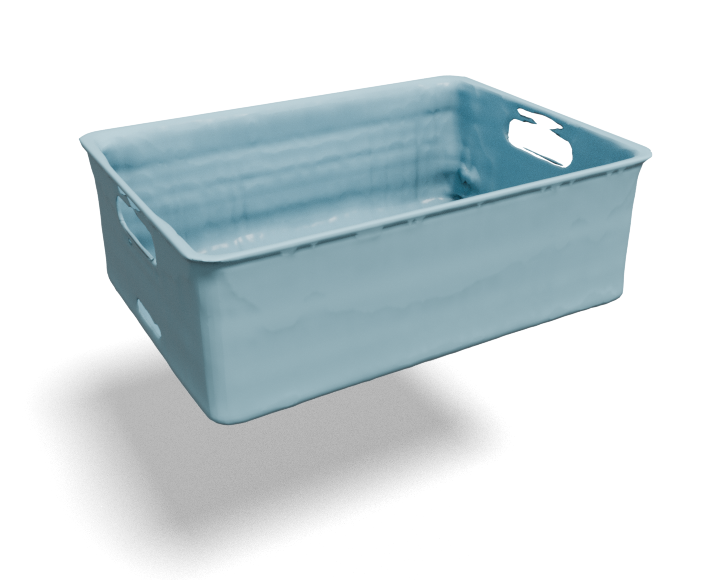}} &  
        {\includegraphics[width=0.145\linewidth]{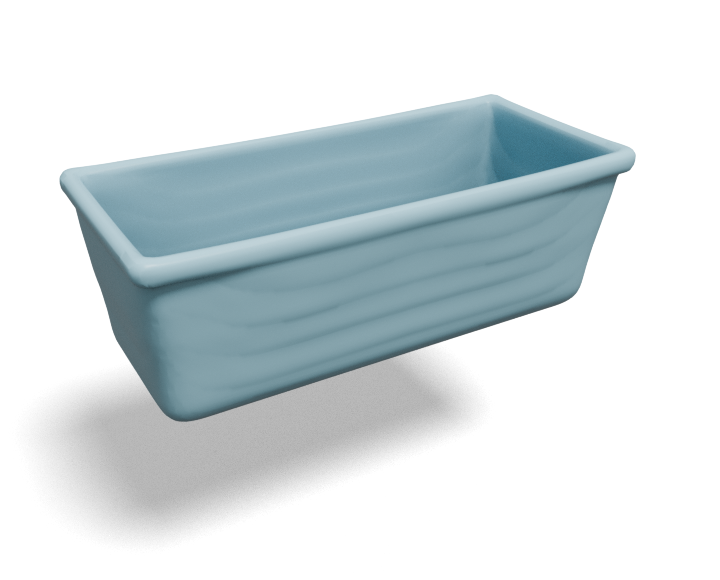}} &
        {\includegraphics[width=0.145\linewidth]{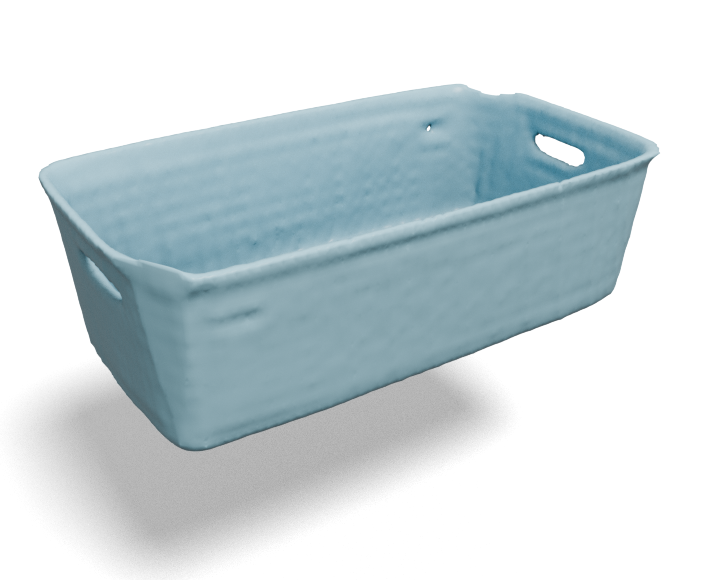}} &
        {\includegraphics[width=0.145\linewidth]{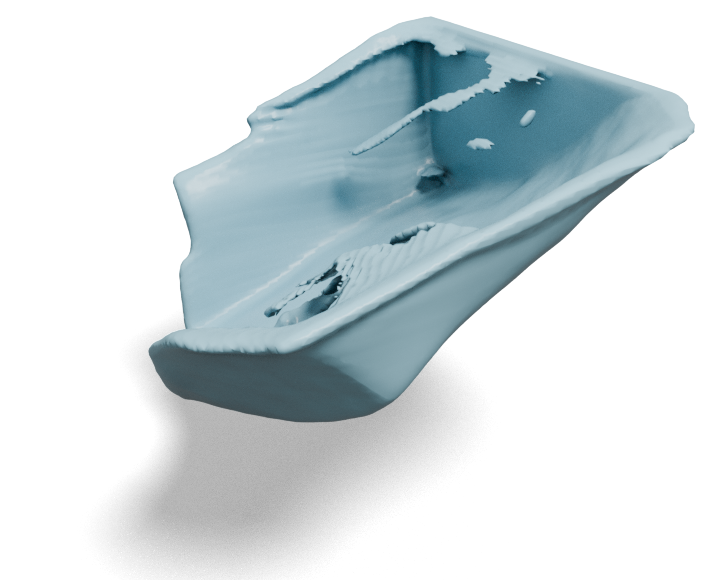}} &
        {\includegraphics[width=0.145\linewidth]{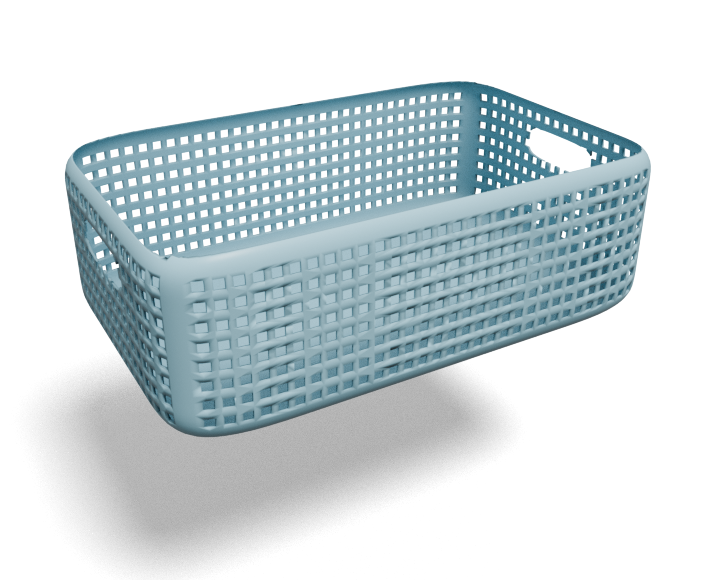}}\\

        {\includegraphics[width=0.145\linewidth]{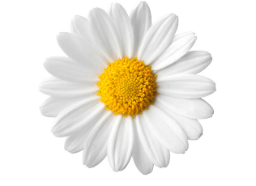}} &  
        {\includegraphics[width=0.145\linewidth]{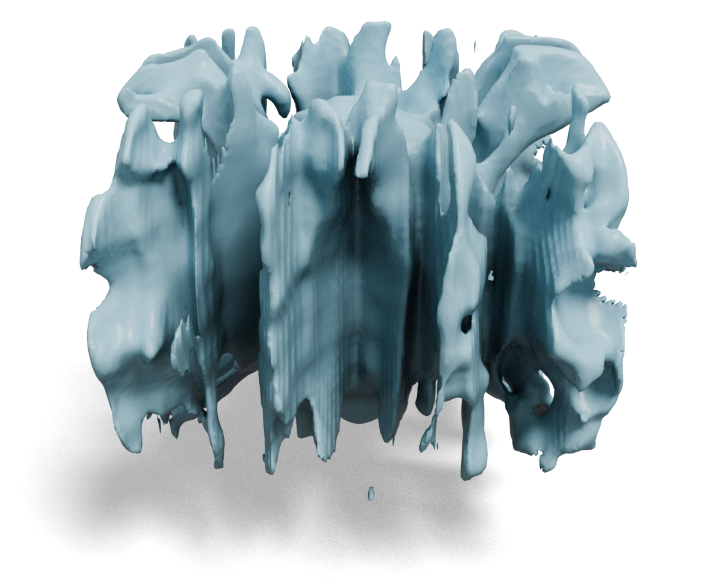}} &  
        {\includegraphics[width=0.145\linewidth]{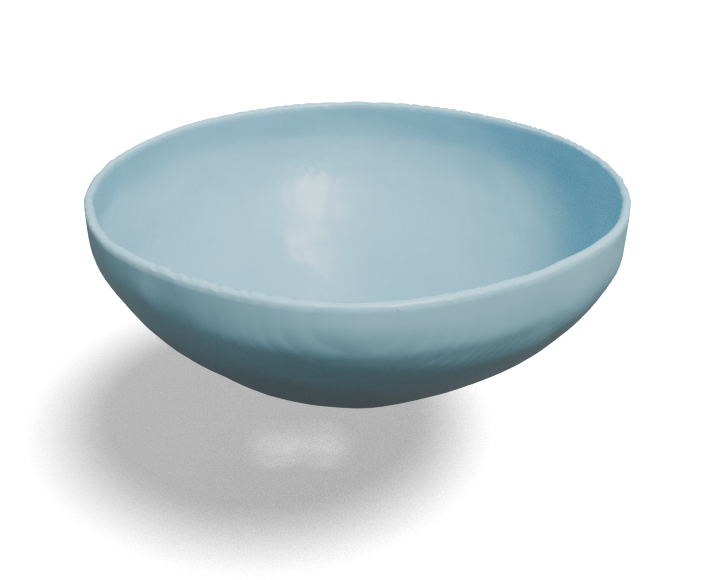}} &
        {\includegraphics[width=0.145\linewidth]{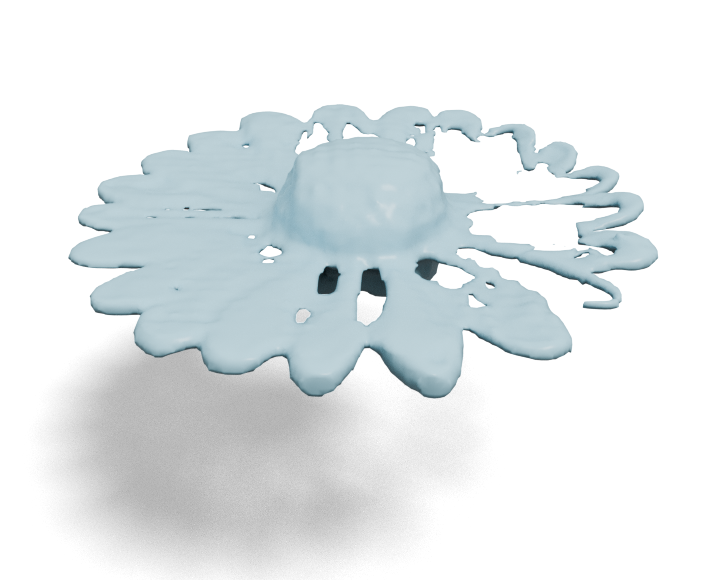}} &
        {\includegraphics[width=0.145\linewidth]{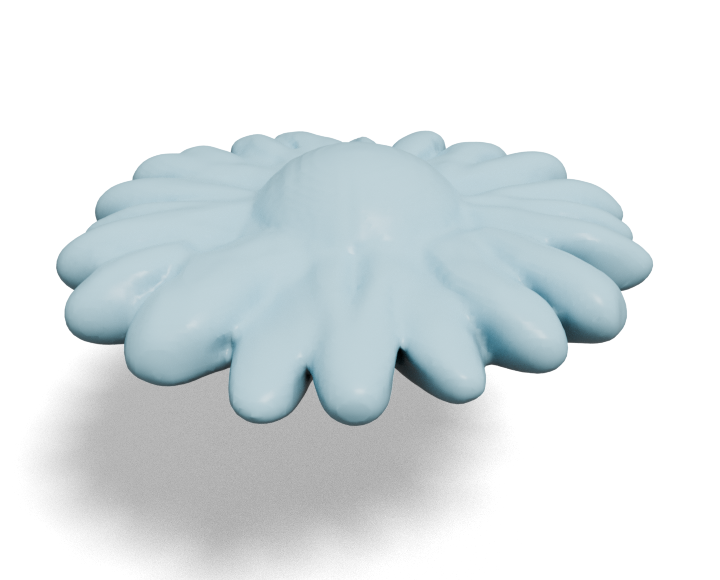}} &
        {\includegraphics[width=0.145\linewidth]{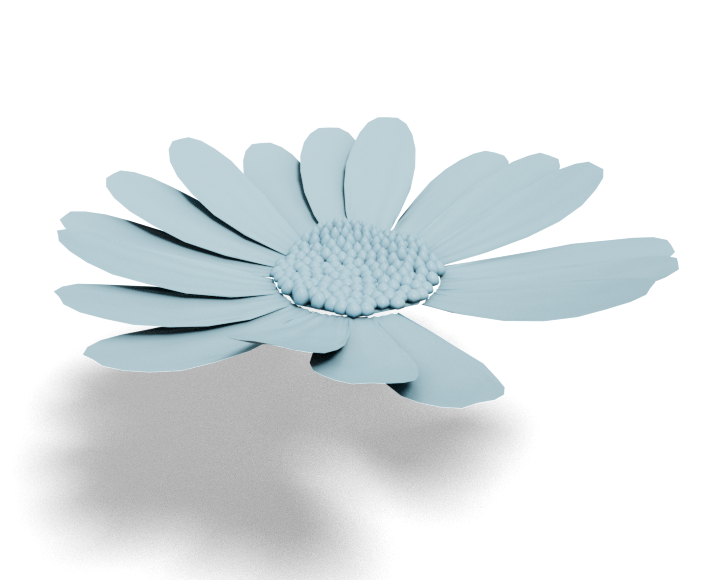}}\\

        {\includegraphics[width=0.145\linewidth]{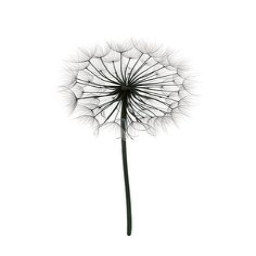}} &  
        {\includegraphics[width=0.145\linewidth]{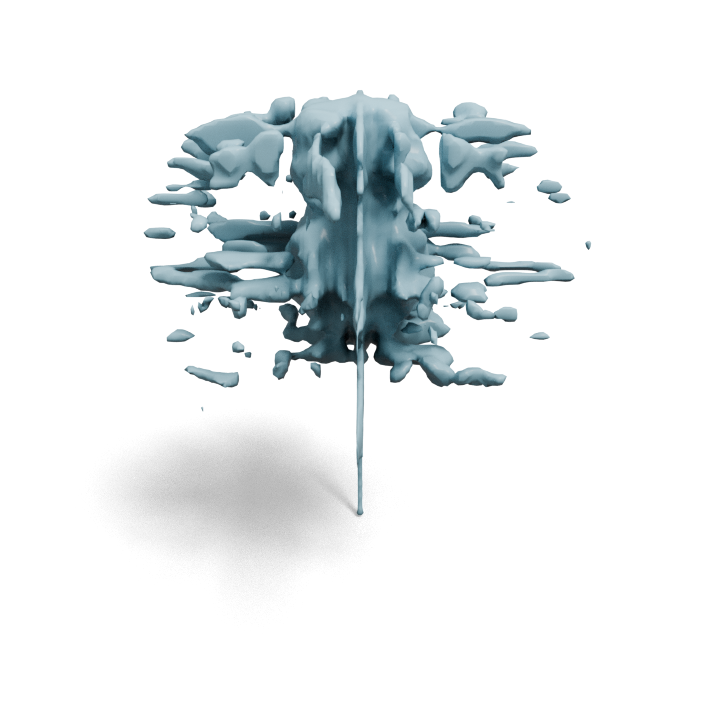}} &  
        {\includegraphics[width=0.145\linewidth]{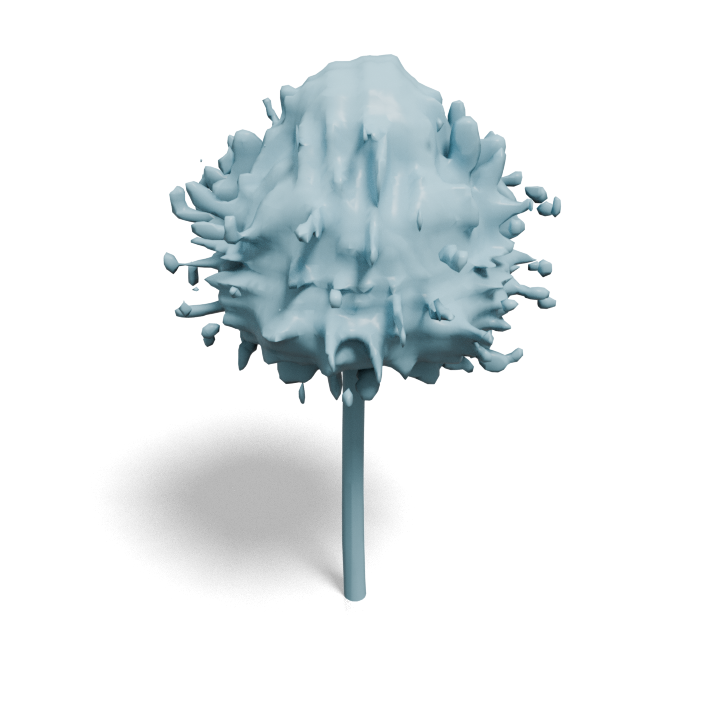}} &
        {\includegraphics[width=0.145\linewidth]{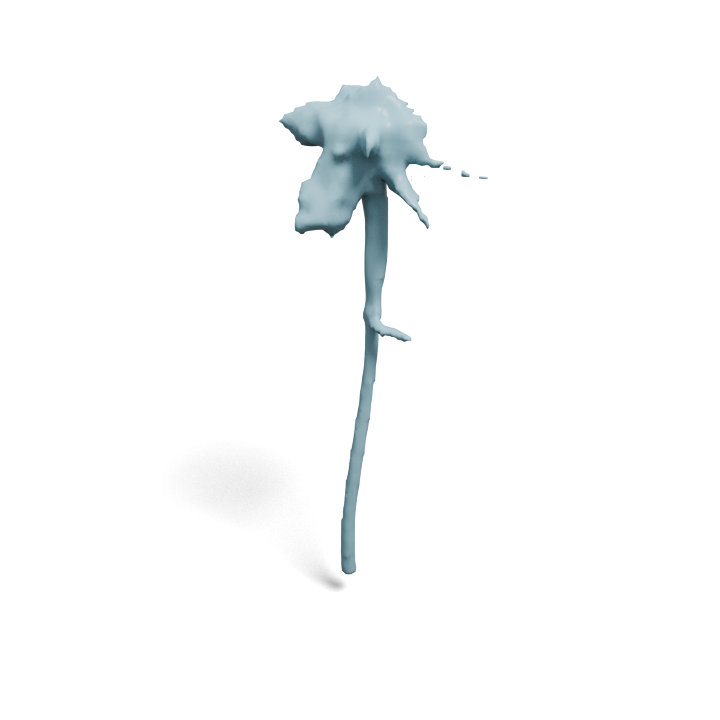}} &
        {\includegraphics[width=0.145\linewidth]{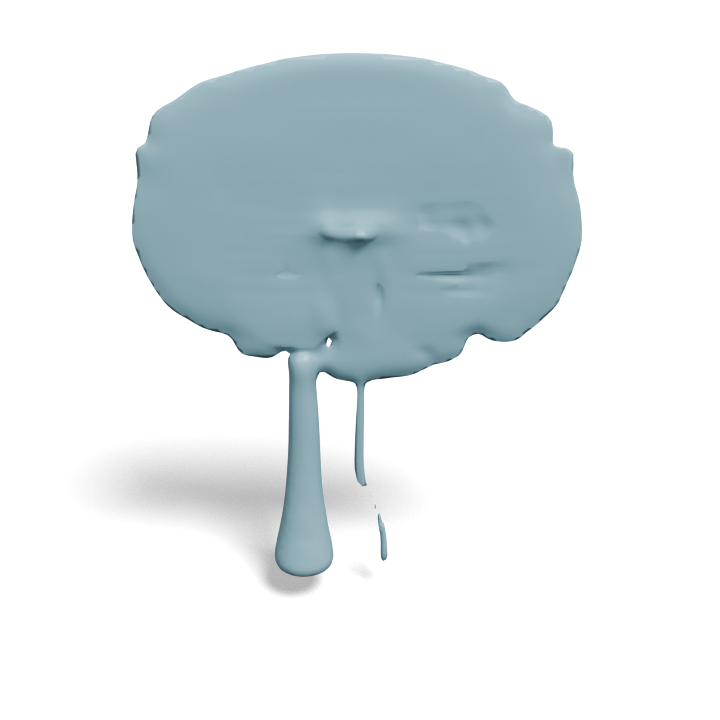}} &
        {\includegraphics[width=0.145\linewidth]{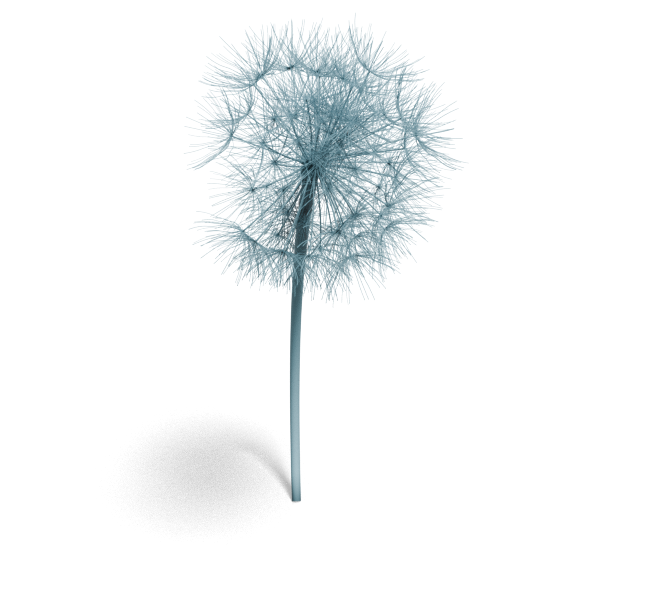}}\\

        Input image & Shap-E~\cite{jun2023shap} & Michelangelo~\cite{zhao2024michelangelo} & InstantMesh~\cite{xu2024instantmesh} & CraftsMan~\cite{li2024craftsman} & DI-PCG \\
        
    \end{tabular}
    \caption{Qualitative comparisons of DI-PCG with baselines.}
    \label{fig::main-comparison}
    \vspace{-10pt}
\end{figure*}
\noindent\textbf{Condition scheme.} DI-PCG takes a single image as the observed data, and injects it into a diffusion model as conditions. To facilitate the generalization ability, we utilize pre-trained visual foundation model to provide general and compact latent representations for images. Specifically, given condition image $I$, we use pre-trained DINOv2~\cite{oquab2023dinov2} model to extract spatial patch features as tokens $\mathbf{c} \in \mathbb{R}^{M\times C}$, where $M$ is the token length and $C$ is the feature channel number. A MLP projector is applied to map the feature tokens to the hidden dimension of DiT. We adopt cross attention to integrate conditions for better spatial alignment, as formulated in Eq.~\ref{eq::dit}.

\noindent\textbf{Data preparation.} DI-PCG is trained with image-parameter pairs generated from the corresponding procedural generator. This self-contained training characteristic is natural and necessary, since the desired objects may be at the end of the long-tail data distribution and hard to collect. To train the diffusion model, we randomly sample parameters, use the procedural generator to build 3D models, and then render RGB images of the model. To improve the generalization ability, multi-view rendering and data augmentation are employed. Specifically, we render the image from the combinations of three azimuths $[0, 30, 60]$, two elevations $[30, 60]$, and two camera distances $[1.8, 2.0]$. Random color augmentation, flipping and cropping are adopted. In addition, we occasionally drop the RGB values and use binary mask as condition, and also sometimes use edge maps from Canny Detector, to enhance the model robustness to texture variations and make it focus on the shapes. 

\noindent\textbf{Implementation details.} To demonstrate the effectiveness of the proposed DI-PCG, we select six procedural generators from Infinigen~\cite{raistrick2023infinite} and Infinigen Indoors~\cite{raistrick2024infinigen}, namely Chair, Table, Vase, Basket, Flower and Dandelion. For each procedural generator, we generate 20000 data pairs following the above mentioned data preparation process, with 18000 for training and 2000 for validation. We train a diffusion model for each procedural generator, resulting in total six diffusion models. These diffusion models have the same model configurations, only except for the input token length, which is determined by the parameter numbers of the generator. Note that DI-PCG is a general method suitable for any procedural generator, without procedural-specific priors in design. Condition images are resized into $256\times256$ resolution and processed by the DINOv2 ViT-B/14 model. Each diffusion model is trained on a single NVIDIA V100 GPU for around 30 hours.
\begin{figure*}[tb]
    \centering
    \small
    \setlength{\tabcolsep}{0pt}
    \begin{tabular}{cccccc}

    \centering

        {\includegraphics[width=0.15\linewidth]{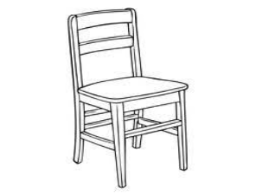}} &  
        {\includegraphics[width=0.15\linewidth]{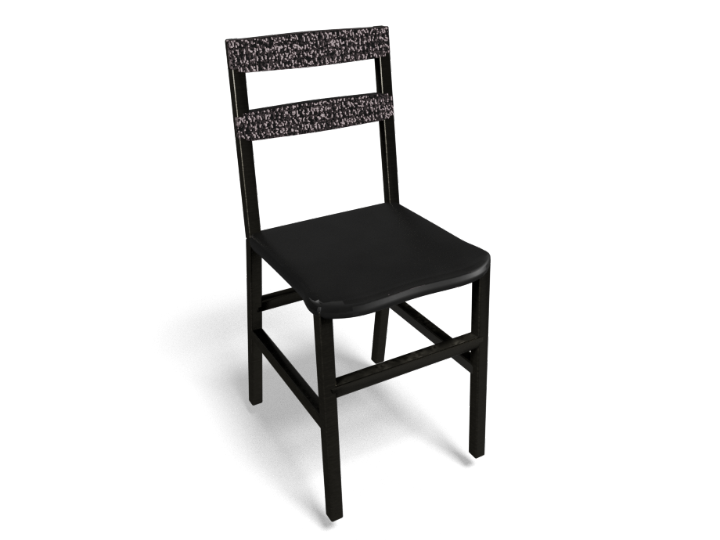}} &  
        {\includegraphics[width=0.15\linewidth]{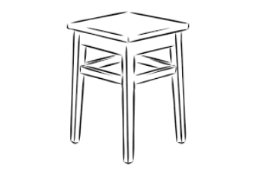}} &
        {\includegraphics[width=0.15\linewidth]{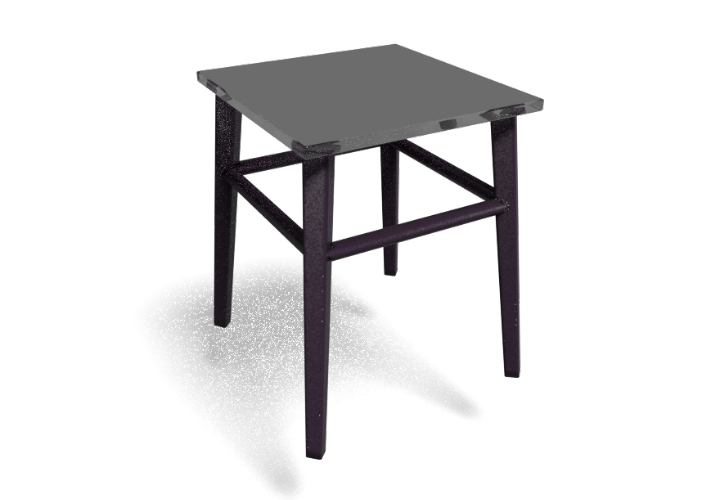}} &
        {\includegraphics[width=0.15\linewidth]{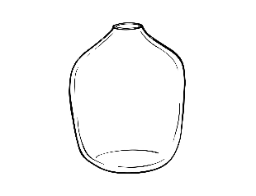}} &
        {\includegraphics[width=0.15\linewidth]{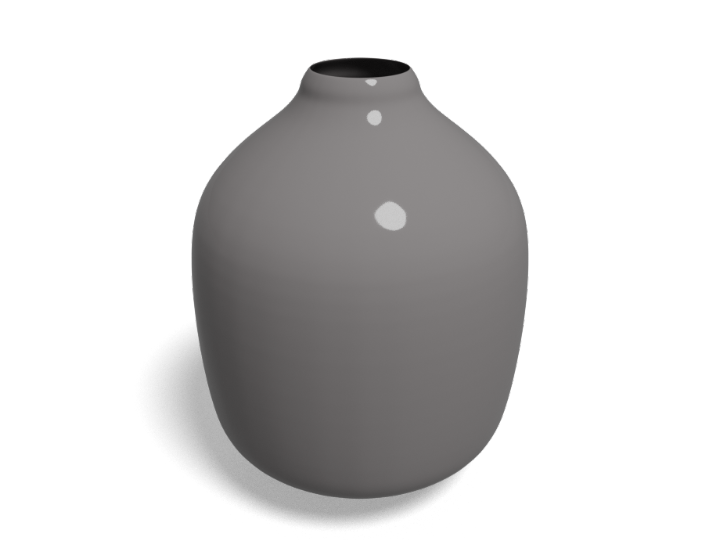}}\\

        {\includegraphics[width=0.15\linewidth]{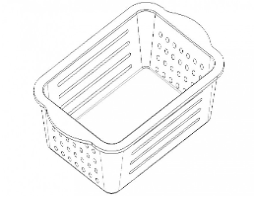}} &  
        {\includegraphics[width=0.15\linewidth]{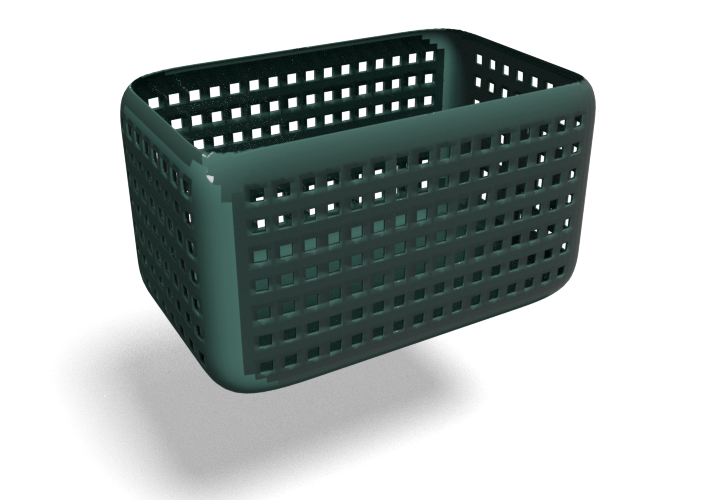}} &  
        {\includegraphics[width=0.15\linewidth]{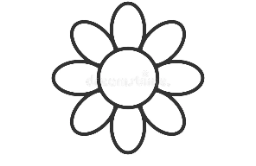}} &
        {\includegraphics[width=0.15\linewidth]{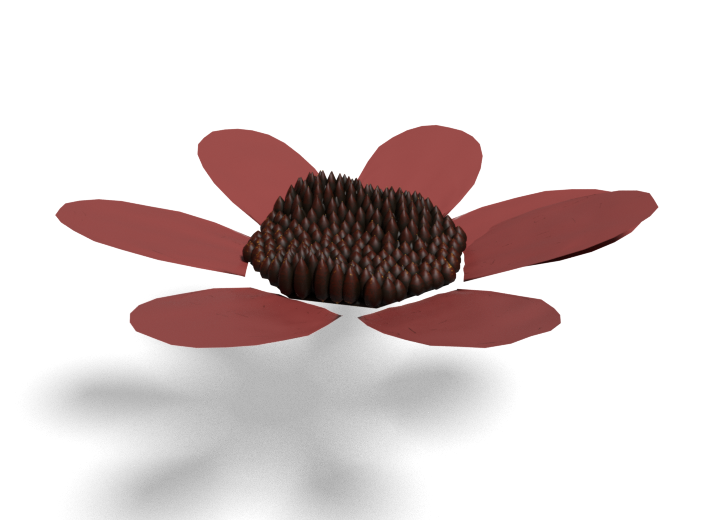}} &
        {\includegraphics[width=0.15\linewidth]{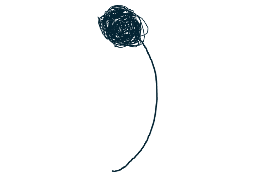}} &
        {\includegraphics[width=0.15\linewidth]{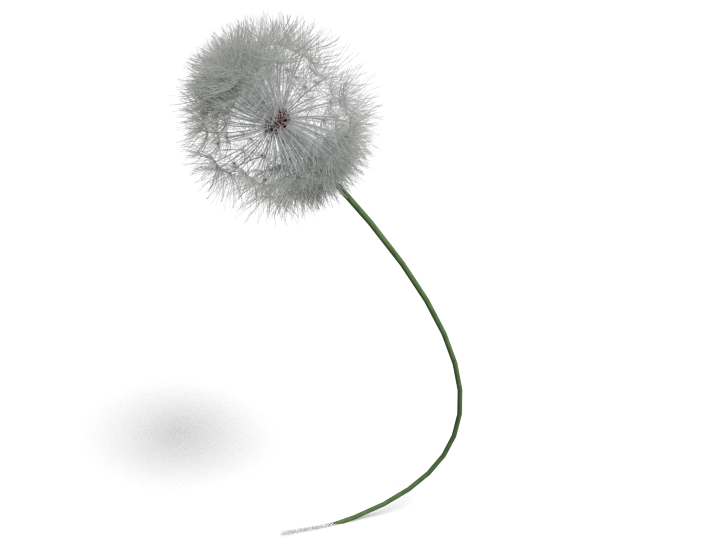}}\\


        Input sketches & Results & Input sketches & Results & Input sketches & Results \\

    \end{tabular}
    \caption{Sketch-conditioned generation results. Textures and materials are randomly picked by the procedural generators.}
    \label{fig::sketch}
    \vspace{-10pt}
\end{figure*}
\begin{table*}[tb]
    \centering
    \begin{tabular}{lccc|ccc}
        \toprule
        \multirow{2}{*}{} & \multicolumn{3}{c}{Test Split of DI-PCG} & \multicolumn{3}{c}
        {ShapeNet Chairs} \\
        \cmidrule(lr){2-4} \cmidrule(lr){5-7}
        & CD$\downarrow$ & EMD$\downarrow$ & F-Score$\uparrow$ & CD$\downarrow$ & EMD$\downarrow$ & F-Score$\uparrow$ \\
        
        \midrule
        Shap-E
        & 0.261 & 0.235 & 0.208 & 0.227 & 0.201 & 0.285\\
        SDFusion
        & 0.252 & 0.234 & 0.167 & 0.255 & 0.244 & 0.178 \\
        Michelangelo
        & 0.181 & 0.171 & 0.289 & 0.111 & 0.125 & 0.407 \\
        CraftsMan
        & 0.253 & 0.231 & 0.189 & 0.177 & 0.168 & 0.280\\
        InstantMesh
        & 0.098 &	0.097 &	0.416 & 0.095 & 0.112 & \textbf{0.473} \\
        DI-PCG & \textbf{0.033} & \textbf{0.028} & \textbf{0.896} & \textbf{0.093} & \textbf{0.108} & 0.452 \\
        \bottomrule
    \end{tabular}
    \caption{Quantitative comparisons on the test split of DI-PCG and the selected ShapeNet chair subset.}
    \label{tab::chair_main}
    \vspace{-10pt}
\end{table*}
\section{Experiments}
To validate the effectiveness of DI-PCG, we conduct detailed experimental evaluations both qualitatively and quantitatively. For baselines, we select representative state-of-the-art 3D reconstruction and generation methods, including 3D native diffusion methods Shap-E~\cite{jun2023shap}, SDFusion~\cite{cheng2023sdfusion}, Michelangelo~\cite{zhao2024michelangelo}, CraftsMan~\cite{li2024craftsman} and large reconstruction model based method InstantMesh~\cite{xu2024instantmesh}.

\subsection{Qualitative Results}
\noindent\textbf{Image condition.} We collect diverse images from internet for all six categories. These images are in multiple styles with different object orientations, delicate geometries, various textures and materials, forming an extensive and challenging test for image-to-3D generation methods. In Figure~\ref{fig::ipcg-results}, we show qualitative results on chair, table, and vase categories. Our method can reliably recover appropriate procedural generator parameters, thus deliver high fidelity 3D generated models of neat geometry, standard meshing and precise alignments with condition images. We recommend readers to supplementary materials for more results. We also conduct comparisons with above mentioned strong baselines. As shown in Figure~\ref{fig::main-comparison}, thanks to its parametric representation upon procedural generators, DI-PCG achieves much improved generation results, being able to preserve intricate details such as holes in basket, dandelion petals, etc. As comparison, Shap-E~\cite{jun2023shap} produces noisy surfaces and fails to handle natural objects like flower and dandelion. Michelangelo~\cite{zhao2024michelangelo} tends to output smooth geometries thanks to its latent representation design, yet lacks sufficient details or misaligned with the image. InstantMesh~\cite{xu2024instantmesh}
and CraftsMan~\cite{li2024craftsman} both rely on multi-view diffusion model to dream about the inputs. While more generalizable than direct 3D methods, they suffer from the inconsistency and errors of the generated multi-view images, and also can not recover complex 3D details.

\noindent\textbf{Sketch condition.} Thanks to the generalization ability of visual foundation model features and our data augmentation strategy, DI-PCG can directly process sketch image conditions and outputs decent 3D generations, as illustrated in Figure~\ref{fig::sketch}. This functionality greatly facilitate the object designs and edits, offering a simple yet effective way to create high-quality 3D assets. More results are included in supplementary materials.

\begin{figure}[tb]
    \centering
    \scriptsize
    \setlength{\tabcolsep}{0pt}
    \begin{tabular}{cccc}

    \centering 
        {\includegraphics[width=0.25\linewidth]{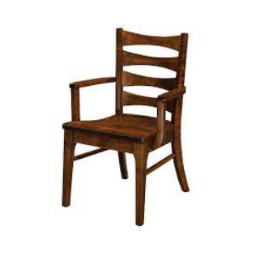}} &
        {\includegraphics[width=0.25\linewidth]{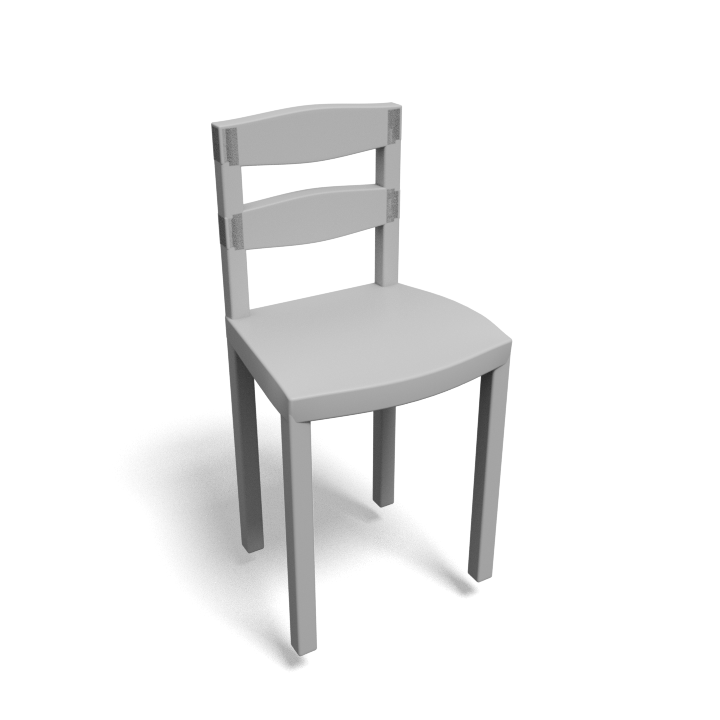}} &
        {\includegraphics[width=0.25\linewidth]{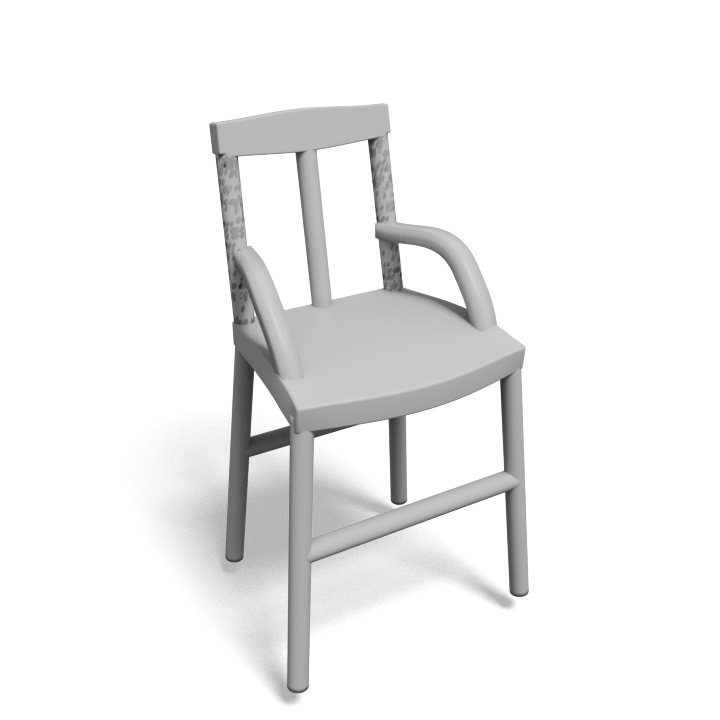}} &
        {\includegraphics[width=0.25\linewidth]{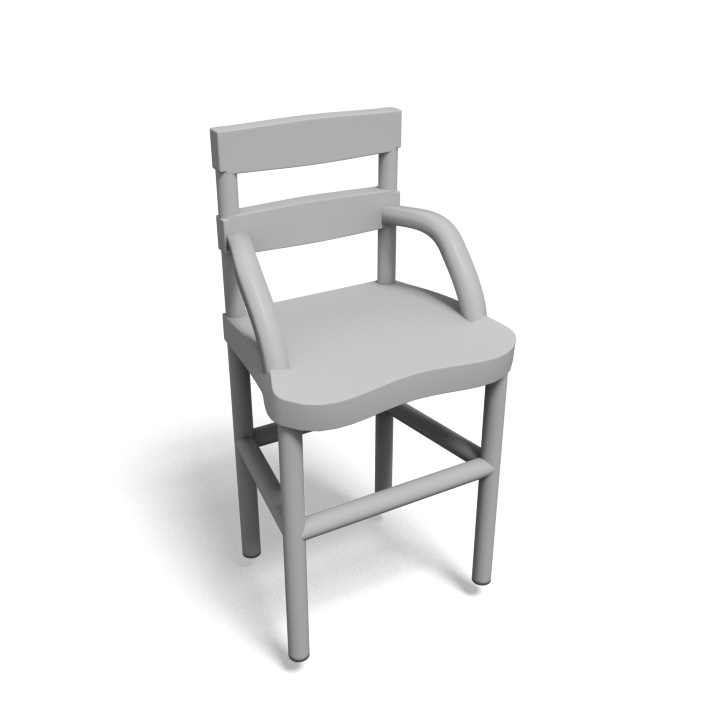}}\\

        Input image & 100 iters / 2 mins & 500 iters / 8 mins & 1k iters / 17 mins\\

        {\includegraphics[width=0.25\linewidth]{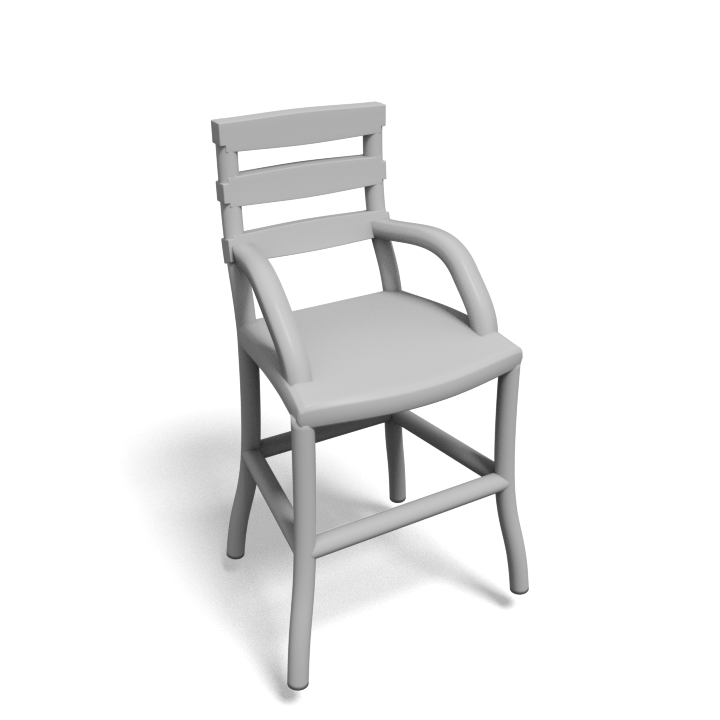}} &
        {\includegraphics[width=0.25\linewidth]{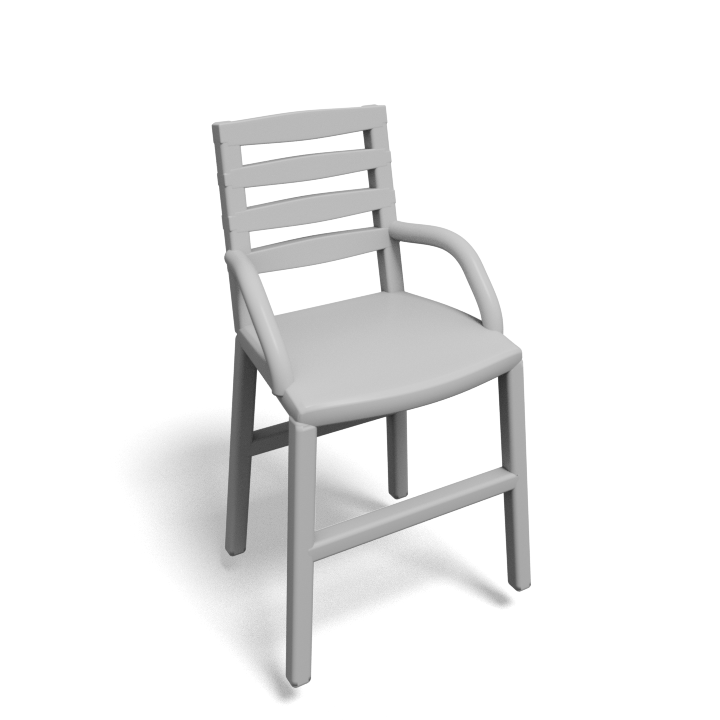}} &
        {\includegraphics[width=0.25\linewidth]{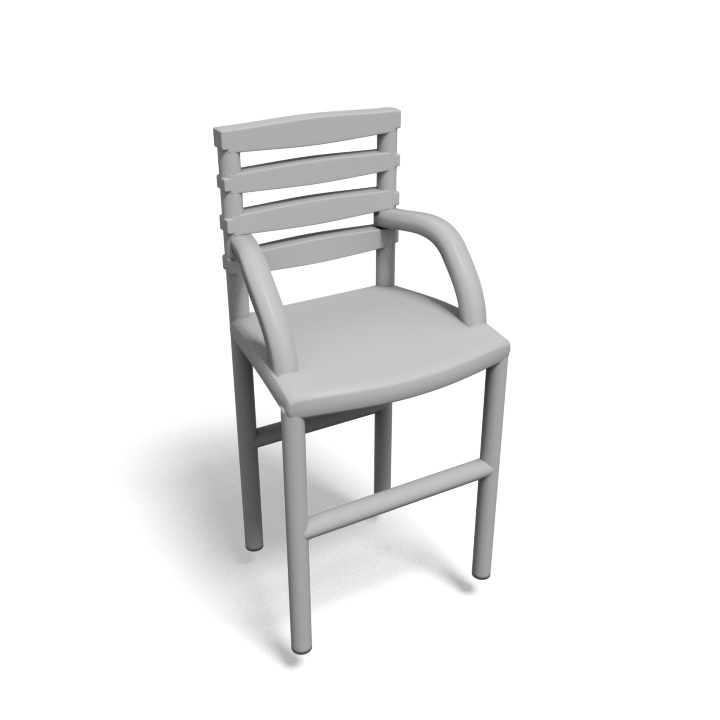}} &
        {\includegraphics[width=0.25\linewidth]{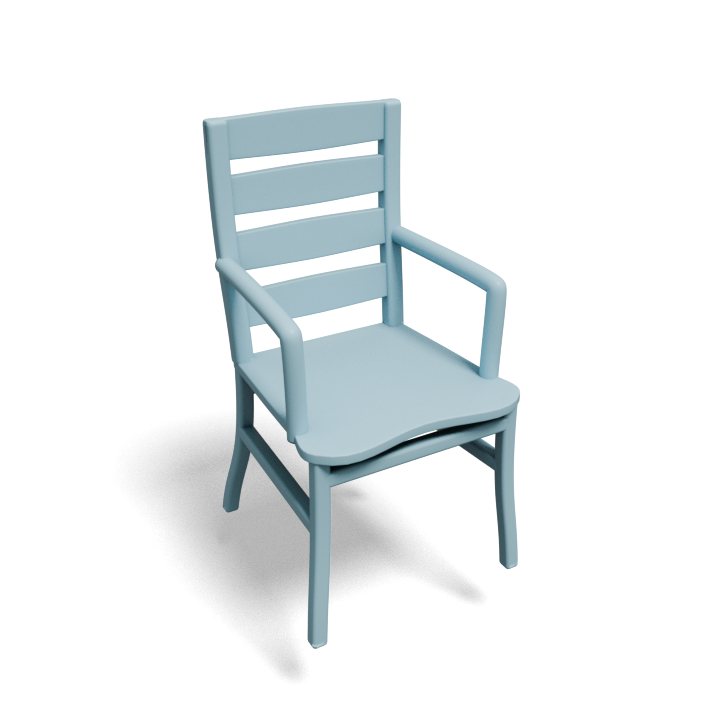}}\\

        2k iters / 33 mins & 5k iters / 83 mins & 10k iters / 167 mins & DI-PCG / 5 secs\\

    \end{tabular}
    \caption{Example of comparison with MCMC method.}
    \label{fig::mcmc}
    \vspace{-10pt}
\end{figure}
\noindent\textbf{Comparison with MCMC.} As the representative sampling method for inverse PCG, MCMC can effectively approximate the parameter distribution, with the presence of powerful scoring metrics and sufficient iterations. We implement a vanilla MCMC method with Metropolis-Hasting algorithm~\cite{metropolis1953equation}, and employ DINOv2~\cite{oquab2023dinov2} as the scorer. 
The DINOv2 scores each sample by calculating the feature distance between input condition image and the rendered image from generated 3D models. As shown in Figure~\ref{fig::mcmc}, the MCMC method outputs gradually closer results as the sampling continues, yet often requires thousands of iterations to complete. Due to the costly forward process of Infinigen generators, it can take several hours. The final results may still contain errors due to the limited ability of the scorer and sensitive hyperparameters. Compared to MCMC, DI-PCG learns the target distribution priors with diffusion models, thus can directly sample the desired parameters with high precision in only several seconds.

\subsection{Quantitative Comparison}
For quantitative evaluations, we use the chair category to demonstrate since it is commonly used and widely available in existing datasets. In addition to the evaluation on test split of our generated data, we also test on the ShapeNet~\cite{chang2015shapenet} chair models to verify its generalization ability. Specifically, we follow the split of 3D-R2N2~\cite{choy20163d}, and manually filter the test chair models to exclude totally out-of-domain samples such as sofa-like or artistic-designed chairs which are currently impossible for Infinigen~\cite{raistrick2023infinite} chair generator to model. The resulting ShapeNet chairs contain 218 models for testing.  We adopt commonly used 3D metrics Chamfer Distance (CD), Earth Moving Distance (EMD), and F-Score.  Table~\ref{tab::chair_main} summarizes the results. It clearly shows that DI-PCG can reliably fit the procedural generator and inversely estimate the parameters with high accuracy. Moreover, it generalizes beyond the procedurally generated chairs and achieves comparable or even better results than previous SOTA methods on ShapeNet chairs subset.

\begin{table}[tb]
    \centering
    \begin{tabular}{lccc}
        \toprule
        & CD$\downarrow$ & EMD$\downarrow$ & F-Score$\uparrow$ \\
        \midrule
        w/o MV \& Aug & 0.139 & 0.140 & 0.321\\
        DI-CLIP & 0.161 & 0.163 & 0.288 \\
        DI-Small (1.6M) & 0.108 & 0.121 & 0.423 \\
        DI-Large (39M) & 0.094 & 0.110 & \textbf{0.452} \\
        \midrule
        DI-PCG & \textbf{0.093} & \textbf{0.108} & \textbf{0.452} \\
        \bottomrule
    \end{tabular}
    \caption{Ablation studies on ShapeNet chairs subset.}
    \label{tab::ablation}
    \vspace{-10pt}
\end{table}
\subsection{Ablation Study}
We conduct ablation studies for different components of DI-PCG. The results are obtained on the above mentioned ShapeNet chair subset, summarized in Table~\ref{tab::ablation}. \textit{w/o MV \& Aug} indicates generating training data with single view image and no augmentations, thus the performance is degraded. \textit{DI-CLIP} denotes using CLIP instead of DINOv2 as the feature for condition. It clearly verifies the effectiveness of DINOv2 features on capturing rich shape features. To study the effect of model size, we train another two diffusion models with small (1.6M parameters) and large (39M parameters) network configurations. As shown in the table, a larger model with more parameters is not necessary and provides no improvements. While small model indeed causes some performance degradation, the trade-off between model size and performance is reasonable and provides more options for different scenarios.

\subsection{Editing Application}
\begin{figure}[tb]
    \centering
    \footnotesize
    \setlength{\tabcolsep}{0pt}
    \begin{tabular}{cccc}

    \centering 
        {\includegraphics[width=0.25\linewidth]{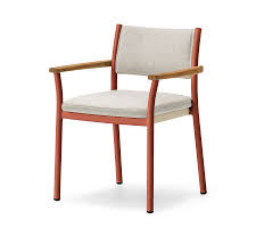}} &
        {\includegraphics[width=0.25\linewidth]{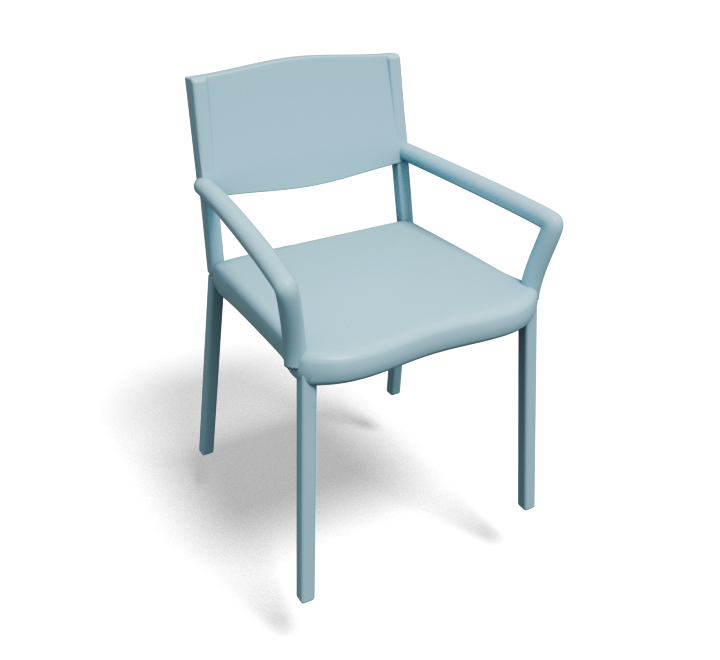}} &
        {\includegraphics[width=0.25\linewidth]{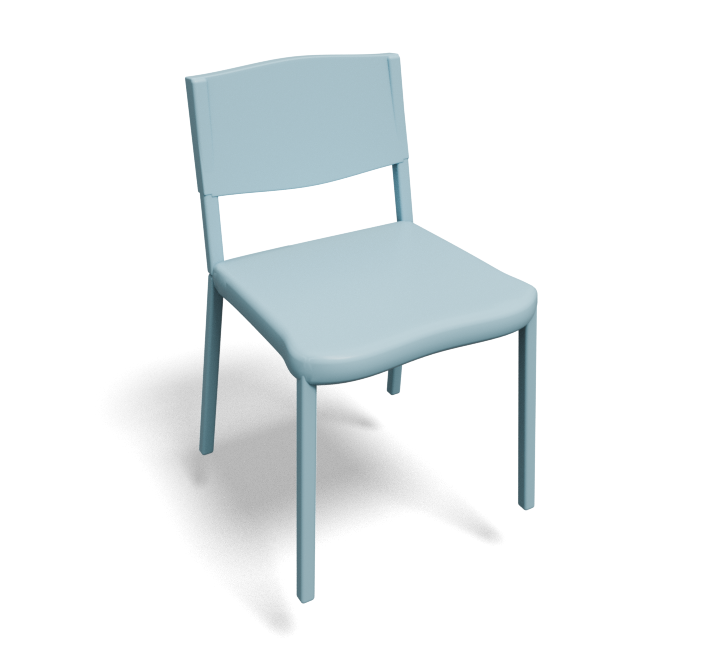}} &
        {\includegraphics[width=0.25\linewidth]{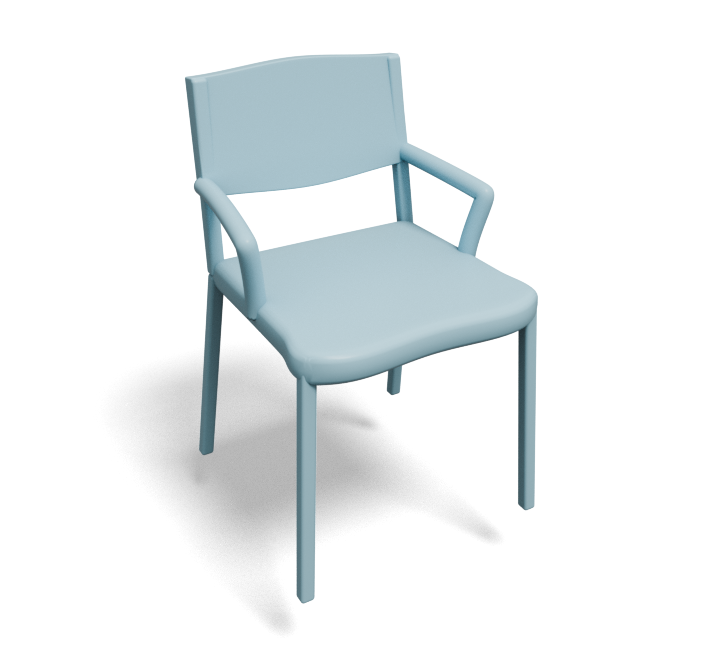}}\\

        Input image & Original & Edit - No arm & Edit - Short arm\\

        {\includegraphics[width=0.25\linewidth]{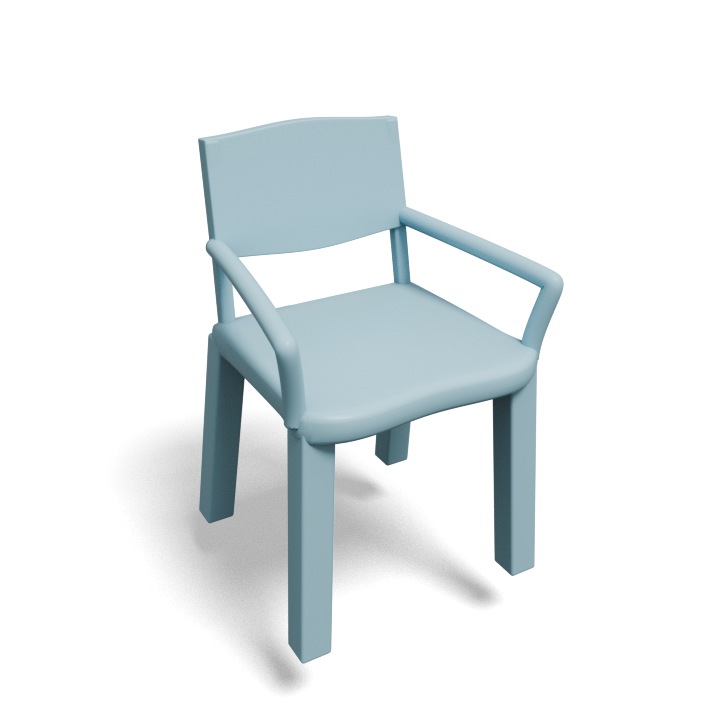}} &
        {\includegraphics[width=0.25\linewidth]{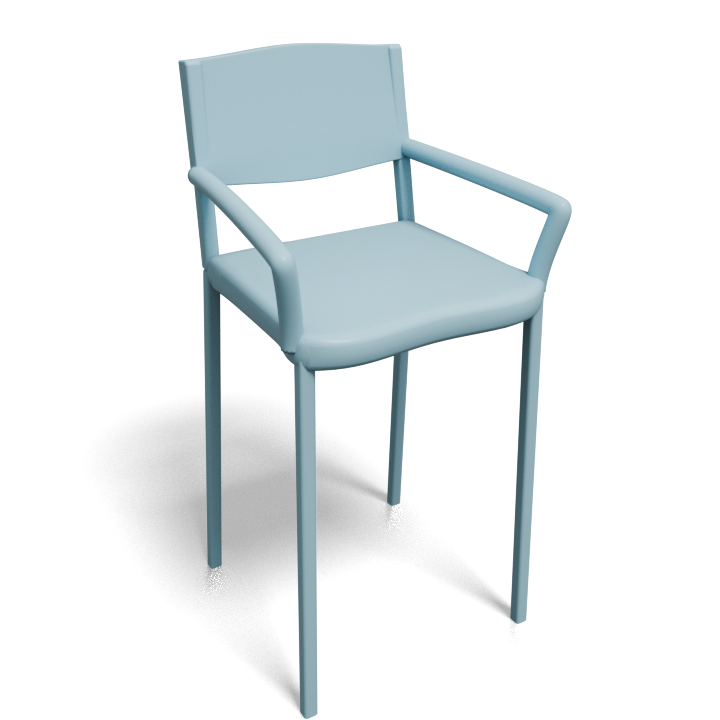}} &
        {\includegraphics[width=0.25\linewidth]{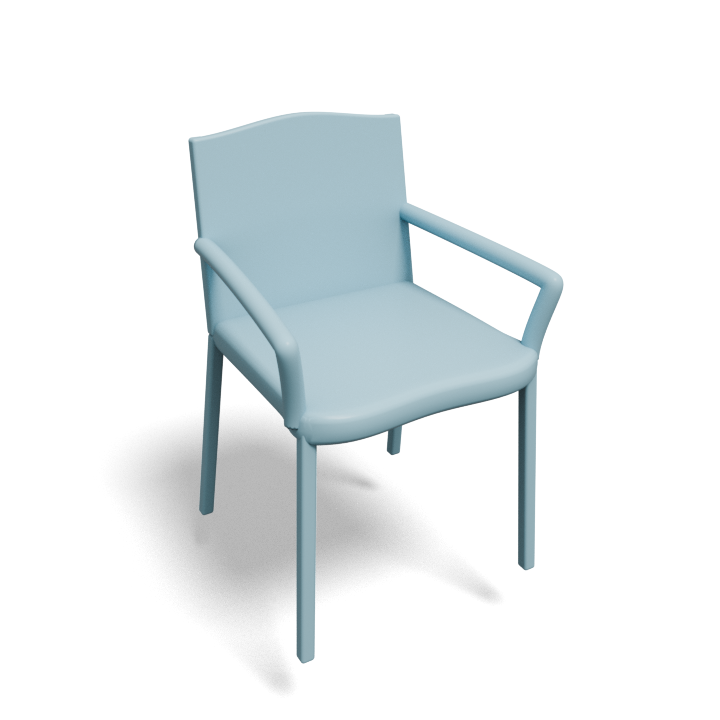}} &
        {\includegraphics[width=0.25\linewidth]{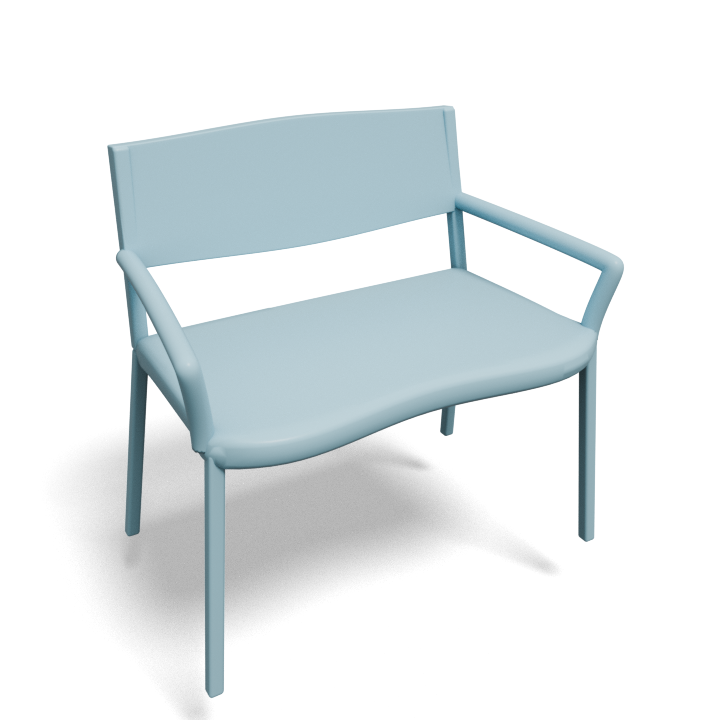}}\\

        Edit - Thick legs & Edit - Taller & Edit - Whole back & Edit - Wider\\

    \end{tabular}
    \caption{DI-PCG supports easy editing by simply adjusting corresponding parameters. }
    \label{fig::edit}
    \vspace{-10pt}
\end{figure}
Thanks to the explicit and semantically meaningful characteristic of the procedural generator parameters, we can easily adjust specific parameter values to edit the 3D model. Some simple editing examples are shown in Figure~\ref{fig::edit}, where the geometric attributes of the given chair, such as leg height, back types, are easily changed. We argue that this handy editing functionality is not in conflict with the controlling difficulty of PCG. It would be painful to find suitable combinations of tens of parameters from scratch, but it is easy and natural to adjust one or two specific parameters to edit existing 3D models. In this way, DI-PCG, as an efficient and effective inverse PCG method, unleashes the controlling advantage of procedural generation.

\subsection{Limitations and Future Works}
As an early attempt to explore diffusion-based inverse PCG for 3D generation, DI-PCG has limitations. First, since DI-PCG relies on off-the-shelf procedural generators, the generation scope is strictly bounded by these generators, i.e. DI-PCG cannot generate out-of-domain objects beyond current generators. Some failure examples in the supplementary materials illustrate this shortage. Second, current DI-PCG only supports image as conditions, while text conditions are widely used in 3D AIGC. Finally, DI-PCG is demonstrated on the object generators, and its applicability on scene-level procedural generation is not verified. Future works include extension to scene generation, more conditions, and automatic generation of procedural generators. 
\section{Conclusion}
In this paper, we present DI-PCG, an innovative diffusion-based efficient inverse procedural content generation method for creating high-quality 3D assets. By directly modeling procedural generator parameters as diffusion denoising variables, the posterior distribution of parameters given condition images can be efficiently determined by the learned diffusion model. DI-PCG solves the inverse PCG problem with high efficiency and accuracy, validated by both quantitative and qualitative evaluations. It represents a valuable exploration towards a promising path for 3D content generation, where parametric models and algorithmic rules together play the roles.

\clearpage
{
    \small
    \bibliographystyle{ieeenat_fullname}
    \bibliography{main.bbl}
}

\clearpage
\setcounter{page}{1}
\maketitlesupplementary

\section{More Implementation Details}
We use six procedural generators from Infinigen and Infinigen Indoors, namely chair, table, vase, basket, flower and dandelion generators. They contain 48, 19, 12, 14, 9, 15 controllable parameters, respectively. These are also the input token lengths of each diffusion models, as the procedural parameters directly serve as the denoising variables. \textbf{Our code will be released once the paper is public}.

\section{More Qualitative Results}
\label{sec::supp_qualitative}
\begin{figure*}[tb]
    \centering
    \centering
        \includegraphics[width=0.95\linewidth]{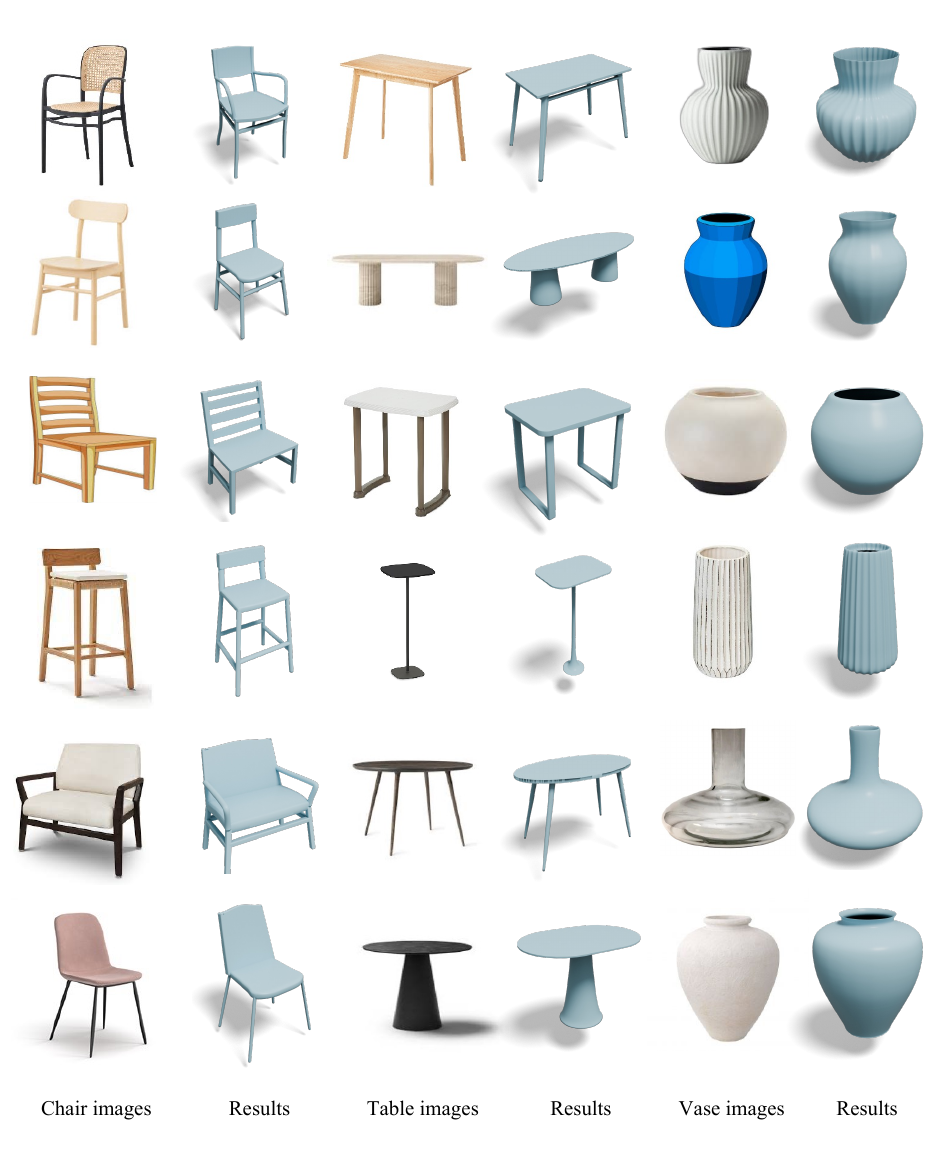}

    \caption{More qualitative results for chair, table, and vase generations. Input images are collected from the internet. DI-PCG can handle diverse input images with various styles, views and textures. It accurately captures geometric details in the input images and generates high fidelity 3D models, facilitating downstream applications.}
    \label{fig::supp-ipcg-results}
\end{figure*}

\begin{figure*}[tb]
    \centering
    \centering
        \includegraphics[width=0.95\linewidth]{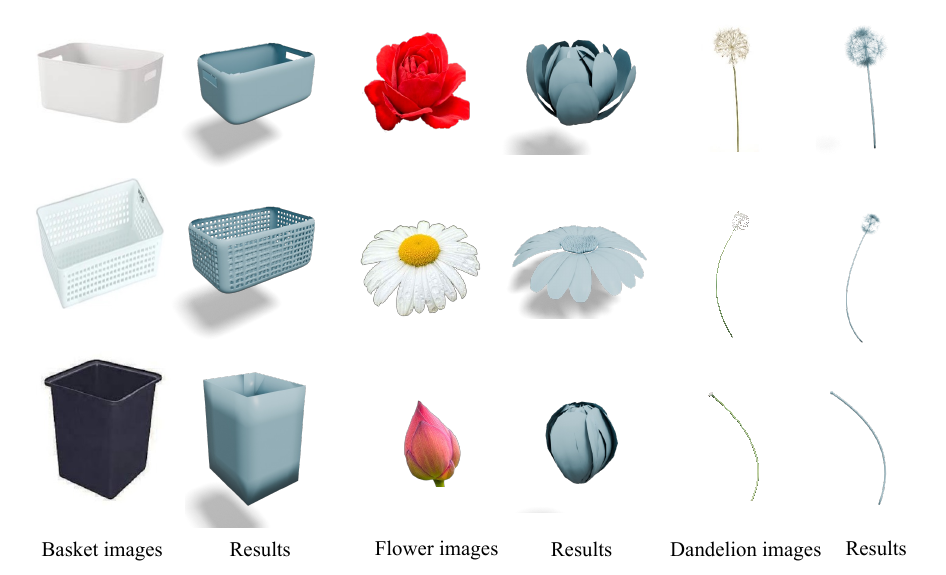}

    \caption{More qualitative results for basket, flower and dandelion generations. }
    \label{fig::supp-ipcg-main2}
\end{figure*}

Here we show more qualitative results of DI-PCG. The generation results for the chair, table, and vase categories are shown in Figure~\ref{fig::supp-ipcg-results}. DI-PCG can handle complex shape variations and details, generating high-quality 3D models from input single images. The results for the basket, flower, and dandelion are shown in Figure~\ref{fig::supp-ipcg-main2}. These categories intrinsically have a bit fewer variations due to the somewhat limited generality of these three procedural generators from Infinigen. Despite that, our method can capture the geometric details and recover the appropriate parameters for the input images, generating fine 3D geometries. 

\begin{figure*}[tb]
    \centering
    \centering
        \includegraphics[width=0.95\linewidth]{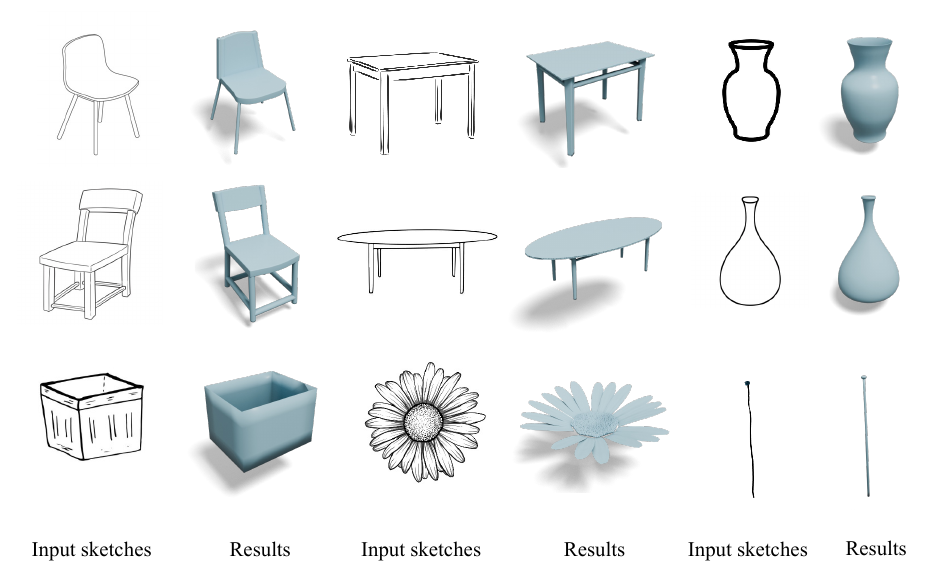}

    \caption{More qualitative results for sketch inputs. DI-PCG can effectively process sketch inputs, offering a convenient way to design and edit objects.}
    \label{fig::supp-ipcg-sketch}
\end{figure*}

DI-PCG can effectively handle sketch input as conditions. We show qualitative examples in Figure~\ref{fig::supp-ipcg-sketch}. In our experiments, we observe that DI-PCG works just as well on sketch inputs as on RGB image inputs. This provides DI-PCG more flexibility and less burden to cooperate with artists.

\begin{figure*}[tb]
    \centering
    \centering
        \includegraphics[width=0.95\linewidth]{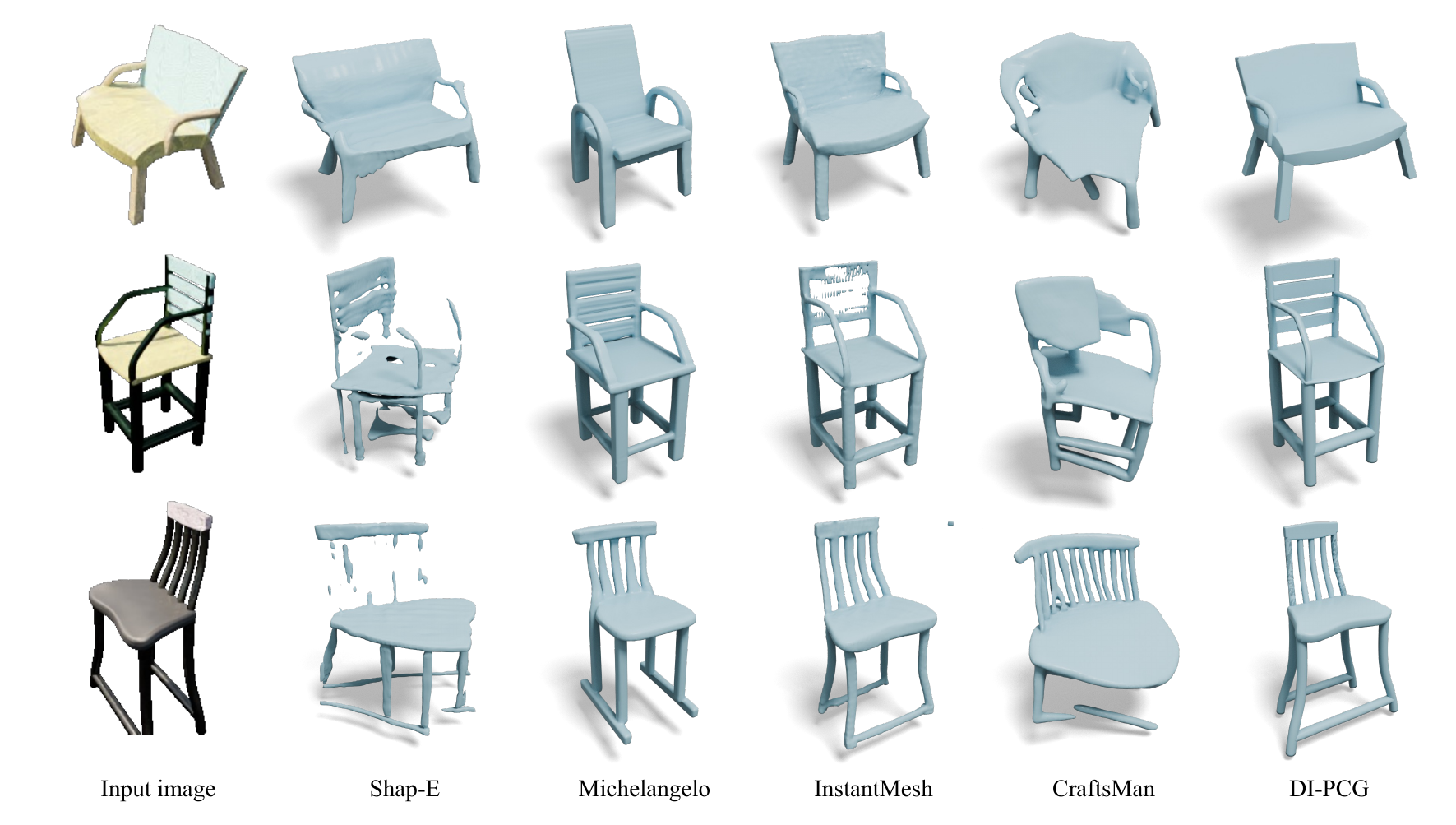}

    \caption{Example comparisons on DI-PCG's test split of chairs. Only DI-PCG generates aligned and clean 3D models.}
    \label{fig::supp-chair-ipcg}
\end{figure*}

\begin{figure*}[tb]
    \centering
    \centering
        \includegraphics[width=0.95\linewidth]{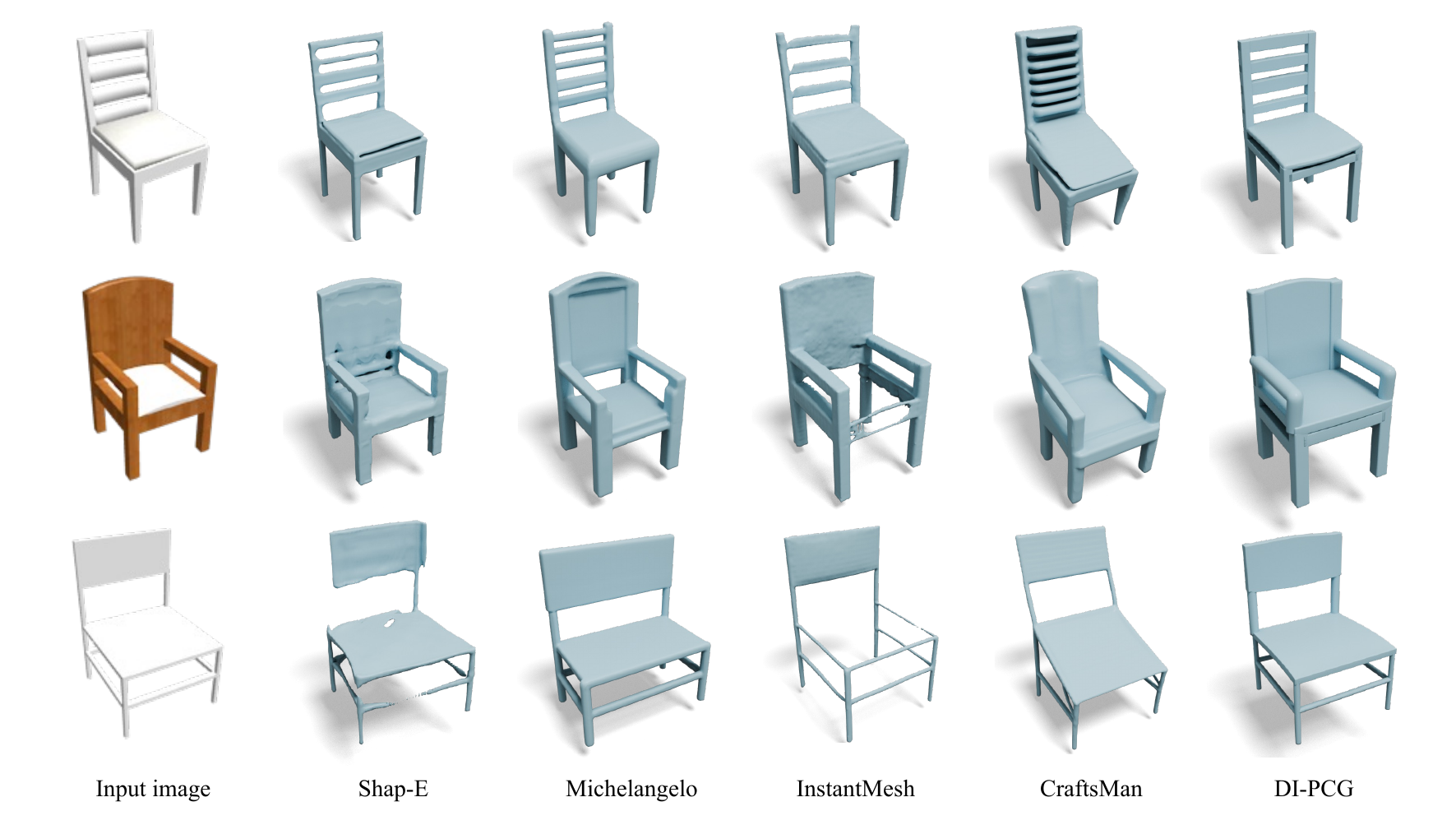}

    \caption{Example comparisons on ShapeNet chair subset. DI-PCG generalizes well and produce high quality 3D meshes.}
    \label{fig::supp-chair-shapenet}
\end{figure*}

We also provide some visual examples of our quantitative evaluations on DI-PCG's test split and ShapeNet chair subset. As shown in Figure~\ref{fig::supp-chair-ipcg} and ~\ref{fig::supp-chair-shapenet}, compared to existing SOTA reconstruction and generation methods, DI-PCG delivers much better 3D models with neat geometry.

\section{Discussions and Failure Cases}
\begin{figure}[tb]
    \centering
    \centering
        \includegraphics[width=0.95\linewidth]{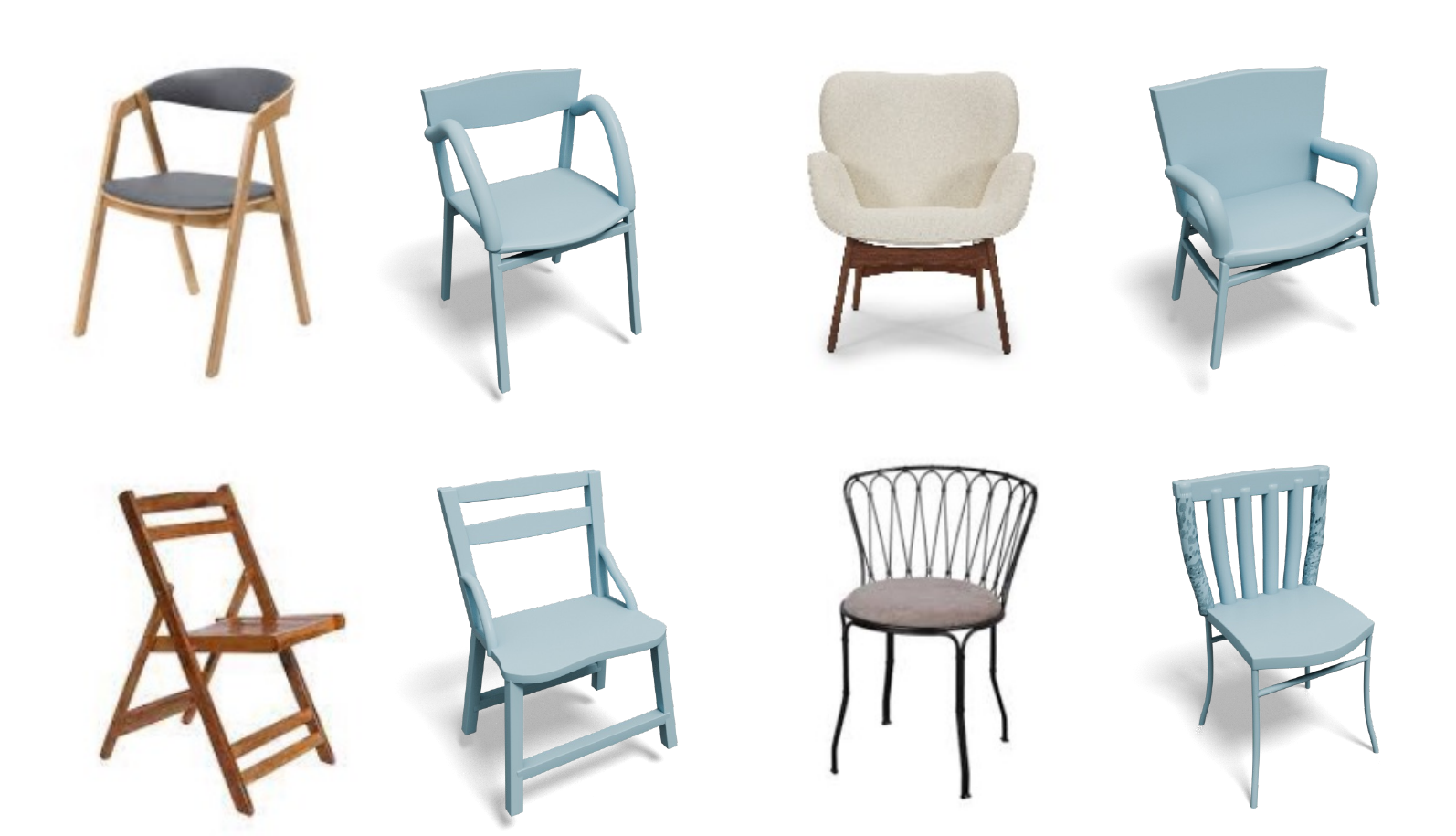}

    \caption{Some failure cases.}
    \label{fig::supp-failure}
\end{figure}

As discussed in the main paper, DI-PCG is limited by the generality and granularity of the given procedural generators. Although the adopted generator from Infinigen can cover a wide range of common variations of the corresponding category, it still has obvious boundaries. Figure~\ref{fig::supp-failure} shows some failure cases. The input chair images are out-of-domain samples for Infinigen chair generators, thus DI-PCG can not generate precisely aligned 3D models. Instead, it outputs the closest parameter sets to approximate the images. Although bounded by the procedural generator, DI-PCG focus on the efficient inverse ability of PCG, and represents a general tool to easily and effectively control any existing procedural generator, facilitating their usage in 3D content creation. As proceudral generators are getting increasing attention and become mature to develop thanks to the modern design softwares, the available number and cover range of existing procedural generators are rapidly growing, which can further benefit DI-PCG. DI-PCG can be applied for any procedural generator, to greatly enhance its controllability. Moreover, in the future, utilizing AI techniques, such as Large Language Model (LLM), to generate procedural generation programs could be possible and exciting. AI-generated procedural generators and DI-PCG can naturally work together, to form a new paradigm of 3D content generation.



\end{document}